\RequirePackage{fix-cm}
\documentclass[twocolumn]{svjour3}          
\smartqed  
\usepackage{graphicx}
\usepackage[ruled,vlined]{algorithm2e}
\usepackage{hhline}
\usepackage{multirow}
\usepackage{pifont}
\usepackage{hyperref}

\newcommand{\cmark}{\ding{51}}%
\newcommand{\xmark}{\ding{55}}%

\newcommand{\G}{{\mathcal G}}

\newcommand{\D}{{\mathcal D}}

\newcommand{\I}{{\mathcal I}}
\newcommand{\A}{{\mathcal A}}
\newcommand{\T}{{\mathcal T}}
\newcommand{\AG}{{\mathcal{AG}}}

\newcommand{\Prop}{{\mathcal V}}

\newcommand{\CL}{{\mathcal{CL}}}

\newcommand{\Ord}{{\mathcal{OR}}}

\mathchardef\mhyphen="2D

\begin{document}

\title{FMAP: Distributed Cooperative Multi-Agent Planning
}


\author{Alejandro~Torre\~no \and Eva~Onaindia \and \' Oscar~Sapena
}


\institute{Alejandro Torre\~no \at
              Departamento de Sistemas Inform\'aticos y Computaci\'on \\
              Universtitat Polit\`ecnica de Val\`encia\\
			  Camino de Vera, s/n, 46022, Valencia, Spain\\
              \email{atorreno@dsic.upv.es}           
           \and
           Eva Onaind\'ia \at
              \email{onaindia@dsic.upv.es}
		   \and
		   \'Oscar Sapena \at
			  \email{osapena@dsic.upv.es}
}

\date{Received: date / Accepted: date}

\maketitle

\begin{abstract}
This paper proposes FMAP (Forward Multi-Agent Planning), a fully-distributed multi-agent planning method that integrates planning and coordination. Although FMAP is specifically aimed at solving problems that require cooperation among agents, the flexibility of the domain-independent planning model allows FMAP to tackle multi-agent planning tasks of any type. In FMAP, agents jointly explore the plan space by building up refinement plans through a complete and flexible forward-chaining partial-order planner. The search is guided by $h_{DTG}$, a novel heuristic function that is based on the concepts of Domain Transition Graph and frontier state and is optimized to evaluate plans in distributed environments. Agents in FMAP apply an advanced privacy model that allows them to adequately keep private information while communicating only the data of the refinement plans that is relevant to each of the participating agents. Experimental results show that FMAP is a general-purpose approach that efficiently solves tightly-coupled domains that have specialized agents and cooperative goals as well as loosely-coupled problems. Specifically, the empirical evaluation shows that FMAP outperforms current MAP systems at solving complex planning tasks that are adapted from the International Planning Competition benchmarks.

\end{abstract}

\section{Introduction}
\label{intro}
%
%
%
%

Multi-agent planning (MAP) introduces a social approach to planning by which multiple intelligent entities work together to solve planning tasks that they are not able to solve by themselves, or to at least accomplish them better by cooperating \cite{deWeerdt09}. MAP places the focus on the collective effort of multiple agents to accomplish tasks by combining their knowledge and capabilities.

The complexity of solving a MAP task directly depends on its typology. In order to illustrate the features of a MAP task, let us introduce a brief application example.

\begin{figure}
\centering
\includegraphics[width=8cm]{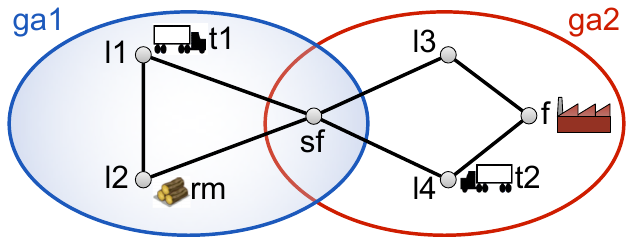}
\caption{Example of a transportation task}
\label{ExampleFig}
\end{figure}

\begin{example} \label{example_task}
Consider the transportation task in Fig. \ref{ExampleFig}, which involves three different agents. There are two transport agencies ($ta1$ and $ta2$), each of which has a truck ($t1$ and $t2$, respectively). The two agencies work in two different geographical areas, $ga1$ and $ga2$, respectively. The third agent is a factory, $f$, which is placed in the area $ga2$. To manufacture products, factory $f$ requires raw materials ($rm$) that are gathered from area $ga1$. In this task, $ta1$ and $ta2$ have the same capabilities, but they act in different areas; i.e., they are spatially distributed agents. Additionally, the factory agent $f$ is functionally different from $ta1$ and $ta2$. The goal of this task is for $f$ to manufacture a set of final products. In order to carry out the task, $ta1$ will send its truck $t1$ to load the raw materials $rm$ located in $l2$ and then transport them to a storage facility ($sf$) that is placed in the intersection of both geographical areas. Then, $ta2$ will complete the delivery by using its truck $t2$ to transport the materials from $sf$ to $f$, which will in turn manufacture the final products. Therefore, this task involves three specialized agents that are spatially and functionally distributed which must cooperate to accomplish a common goal.
\end{example}

Example \ref{example_task} emphasizes most of the key elements of a MAP task. First, the spatial and/or functional distribution of planning agents gives rise to \emph{specialized agents} that have different knowledge and capabilities. In turn, this information distribution stresses the issue of \emph{privacy}, which is one of the basic aspects that should be considered in multi-agent applications \cite{Such13}.

Since the three parties involved in Example \ref{example_task} are specialized in different functional or geographical areas of the task, most of the information managed by factory $f$ is not relevant for the transport agencies and vice-versa. The same occurs with the transport agencies $ta1$ and $ta2$. Additionally, agents may not be willing to share the sensitive information of their internal procedures with the others. For instance, $ta1$ and $ta2$ are cooperating in this particular delivery task, but they might be potential competitors since they work in the same business sector. Therefore, agents in a MAP context want to minimize the information they share with each other, either for strategic reasons or simply because it is not relevant for the rest of the agents in order to address the planning task.

Besides the need for computational or information distribution, privacy is also one of the reasons to adopt a multi-agent approach. This aspect, however, has been traditionally relegated in MAP, particularly by the planning community \cite{Krogt09}. While some approaches define a basic notion of privacy \cite{Borrajo13,Nissim10}, others allow agents to share detailed parts of their plans or do not take private information into account at all \cite{Kvarnstrom11}.

The complexity of a MAP task is often described by means of its \emph{coupling level} \cite{Brafman08}, which is measured as the number of interactions that arise among agents during the resolution of a MAP task. According to this parameter, MAP tasks can be classified into \emph{loosely-coupled tasks} (which present few interactions among agents) and \emph{tightly-coupled tasks} (which involve many interactions among agents). The coupling level, however, does not take into consideration one key aspect of MAP tasks: the presence of \emph{cooperative goals}; i.e., goals that cannot be solved individually by any agent since they require the cooperation of specialized agents. Example \ref{example_task} illustrates a tightly-coupled task with one such goal since none of the agents can achieve the manufacturing of the final products by itself. Instead, they must make use of their specialized capabilities and interact with each other to deliver the raw materials and manufacture the final products.

In this paper, we present FMAP (Forward MAP), which is a domain-independent MAP system that is designed to cope with a great variety of planning tasks of different complexity and coupling level. FMAP is a fully distributed method that interleaves planning and coordination by following a cooperative refinement planning approach. This search scheme allows us to efficiently coordinate agents' actions in any type of planning task (either loosely-coupled or tightly-coupled) as well as to handle cooperative goals.

FMAP relies on a theoretical model which defines a more sophisticated notion of privacy than most of the existing MAP systems. Instead of using a single set of private data, FMAP allows agents to declare the information they will share with each other. For instance, the transport agency $ta2$ in Example 1 will share with factory $f$ information that is likely to be different from the information shared with agent $ta1$. Our system enhances privacy by minimizing the information that agents need to disclose. FMAP is a complete and reliable planning system that has proven to be very competitive when compared to other state-of-the-art MAP systems. The experimental results will show that FMAP is particularly effective for solving tightly-coupled MAP problems with cooperative goals. 

This article is organized as follows: section \ref{related} presents some related work on multi-agent planning, with an emphasis on issues like the coupling level of planning tasks, privacy, or cooperative goals. Section \ref{formalization} formalizes the notion of a MAP task; section \ref{algorithms} describes the main components of FMAP, the search procedure, and the DTG-based heuristic function; finally, section \ref{results} provides a thorough experimental evaluation of FMAP and section \ref{conclusions} concludes the paper.

\section{Related work}
\label{related}

In the literature, there are two main approaches for solving MAP tasks like the one described in Example \ref{example_task}. \emph{Centralized} MAP involves using an intermediary agent that has complete knowledge of the task. The \emph{distributed} or decentralized approach spreads the planning responsability among agents, which are in charge of interacting with each other to coordinate their local solutions, if necessary \cite{Pal13,Kala14}. The adoption of a centralized approach is aimed at improving the planner performance by taking advantage of the inherent structure of the MAP tasks \cite{Kvarnstrom11,Crosby13}. Centralized approaches assume a single planning entity that has complete knowledge of the task, which is rather unrealistic if the parties involved in the task have sensitive private information that they are not willing to disclose \cite{Sapena08}. In Example \ref{example_task}, the three agents involved in the task want to protect the information regarding their internal processes and business strategies, so a centralized setting is not an acceptable solution.

We then focus on fully distributed MAP, that is, the problem of coordinating agents in a shared environment where information is distributed. The distributed MAP setting involves two main tasks: the \emph{planning} of local solutions and the \emph{coordination} of the agents' plans into a global solution. Coordination can be performed at one or various stages of the distributed resolution of a MAP task. Some techniques are used for problems in which agents build local plans for the individual goals that they have been assigned. MAP is about coordinating the local plans of agents so as to mutually benefit by avoiding the duplication of effort. In this case, the goal is not to build a joint plan among entities that are functionally or spatially distributed but rather to apply \emph{plan merging} to coordinate the local plans of multiple agents that are capable of achieving the problem goals by themselves \cite{Cox09}. 

There is a large body of work on plan-merging techniques. The work in \cite{Cox09} introduces a distributed coordination framework based on partial-order planning that addresses the interactions that emerge between the agents' local plans. This framework, however, does not consider privacy. The proposal in \cite{ToninoBWW02} is based on the iterative revision of the agents' local plans. Agents in this model cooperate by mutually adapting their local plans, with a focus on improving their common or individual benefit. This approach also ignores privacy and agents are assumed to be fully cooperative. The approach in \cite{Krogt05b} uses multi-agent plan repair to solve inconsistencies among the agents' local plans while maintaining privacy. $\mu$-SATPLAN \cite{Dimopoulos12} extends a satisfiability-based planner to coordinate the agents' local plans by studying positive and negative interactions among them. 

Plan-merging techniques are not very well suited for coping with tightly-coupled tasks as they may introduce exponentially many ordering constraints in problems that require great coordination effort \cite{Cox09}. In general, plan merging is not an effective method for attaining cooperative goals since this resolution scheme generally assumes that each agent is able to solve a subset of the task's goals by itself. However, some approaches use plan merging to coordinate local plans of specialized agents. In this case, the effort is placed on discovering the interaction points among agents through the public information that they share. For instance, Planning First \cite{Nissim10} introduces a cooperative MAP approach for loosely-coupled tasks, in which specialized agents carry out planning individually through a state-based planner. The resulting local plans are then coordinated by solving a distributed Constraint Satisfaction Problem (CSP) \cite{Jannach13}. This combination of CSP and planning to solve MAP tasks was originally introduced by the MA-STRIPS framework \cite{Brafman08}.

Another major research trend in MAP interleaves planning and coordination, providing a more unified vision of cooperative MAP. One of the first approaches to domain-independent MAP is the Generalized Partial Global Planning (GPGP) framework \cite{Lesser04}. Agents in GPGP have a partial view of the world and communicate their local plans to the rest of the agents, which in turn merge this information into their own partial global plan in order to improve it. Approaches to continual planning (interleaving planning and execution in a world undergoing continual change), assume there is uncertainty in the world state and therefore agents do not have a complete view of the world \cite{Brenner09}. Specifically in \cite{Brenner09}, agents have a limited knowledge of the environment and limited capabilities, but the authors do not explicitly deal with a functional distribution among agents or cooperative goals. TFPOP is a fully centralized approach that combines temporal and forward-chaining partial-order planning to solve loosely-coupled MAP tasks \cite{Kvarnstrom11}. The Best-Response Planning algorithm departs from an initial joint plan that is built through the Planning First MAP system \cite{Nissim10} and iteratively improves the quality of this initial plan by applying cost optimal planning \cite{JonssonR11}. Agents can only access the public information of the other agents' plans thereby preserving privacy, and they optimize their plans with the aim to converge to a Nash equilibrium regarding their preferences. MAP-POP is a fully distributed method that effectively maintains the agents' privacy \cite{Torreno12KAIS,Torreno12ECAI}. Agents in MAP-POP perform an incomplete partial-order planning search to progressively develop and coordinate a joint plan until its completion. 

Finally, MAPR is a recent planner that performs goal allocation to each agent \cite{Borrajo13}. Agents iteratively solve the assigned goals by extending the plan of the previous agent. In this approach, agents work under limited knowledge of the environment by obfuscating the private information in their plans. MAPR is particularly effective for loosely-coupled problems, but it cannot deal with tasks that feature specialized agents and cooperative goals since it assumes that each goal is achieved by a single agent. Section \ref{results} will show a comparative performance evaluation between MAPR and FMAP, our proposed approach.

\section{MAP task formalization}
\label{formalization}

Agents in FMAP work with limited knowledge of the planning task by assuming that information that is not represented in an agent's model is unknown to the agent. The states of the world are modeled through a finite set of \emph{state variables}, $\Prop$, each of which is associated to a finite domain, $\D_v$, of mutually exclusive values that refer to the objects in the world. Assigning a value $d$ to a variable $v \in \Prop$ generates a \emph{fluent}. A \emph{positive fluent} is a tuple $\langle v, d\rangle$, which indicates that the variable $v$ takes the value $d$. A \emph{negative fluent} is of the form $\langle v, \neg d\rangle$, indicating that $v$ does not take the value $d$.  A \emph{state} $S$ is a set of positive and negative fluents. 

An \emph{action} is a tuple $\alpha = \langle PRE(\alpha)$, $EFF(\alpha) \rangle$, where $PRE(\alpha)$ is a finite set of fluents that represents the preconditions of $\alpha$, and $EFF(\alpha)$ is a finite set of positive and negative \emph{variable assignments} that model the effects of $\alpha$. Executing an action $\alpha$ in a world state $S$ leads to a new world state $S'$ as a result of applying $EFF(\alpha)$ over $S$. An effect of the form $(v = d)$ assigns the value $d$ to the variable $v$, i.e., it adds the fluent $\langle v, d\rangle$ to $S'$ as well as adding a set of fluents $\langle v, \neg d'\rangle$ for each other value $d'$ in the variable domain in order to have a consistent state representation. Additionally, any fluent in $S$ of the form $\langle v, \neg d\rangle$ or $\langle v,d''\rangle$, $d'' \neq d$, is removed in state $S'$. This latter modification removes any fluent that contradicts $\langle v, d\rangle$. On the other hand, an assignment $(v \not= d)$ adds the fluent $\langle v, \neg d\rangle$ to $S'$ and removes $\langle v, d\rangle$ from $S'$, if such a fluent exists in $S$. 

For instance, let us suppose that the transportation task in Example \ref{example_task} includes a variable $pos\mhyphen rm$ that describes the position of the raw materials $rm$, which can be any of the locations in the task. Let $S$ be a state that includes a fluent $\langle pos\mhyphen rm, l2\rangle$, which indicates that $rm$ is placed in its initial location (see Fig. \ref{ExampleFig}). Agent $ta1$ performs an action to load $rm$ into its truck $t1$, which includes an effect of the form $(pos\mhyphen rm = t1)$. The application of this action results in a new world state $S'$ that will include a fluent $\langle pos\mhyphen rm, t1\rangle$ and fluents of the form $\langle pos\mhyphen rm, \neg l\rangle$ for each other location $l \neq t1$; the fluent $\langle pos\mhyphen rm, l2\rangle$ will no longer be in $S'$.

\vspace{0,1cm}

\begin{definition}
A \textbf{MAP task} is defined as a tuple $\T_{MAP} = \langle\AG,$$\Prop,$$\I,\G,\A$$\rangle$. $\AG=\{1, \ldots, n\}$ is a finite non-empty set of agents. $\Prop=\bigcup_{i \in \AG}\Prop^i$, where $\Prop^i$ is the set of state variables known to an agent $i$. $\I=\bigcup_{i \in \AG}\I^i$ is a set of fluents that defines the initial state of $\T_{MAP}$. Since specialized agents are allowed, they may only know a subset of $\I$. Given two agents $i$ and $j$, $\I^i \cap \I^j$ may or may not be $\emptyset$; in any case, the initial states of the agents never contradict each other. $\G$ is the set of goals of $\T_{MAP}$, i.e., the values of the state variables that agents have to achieve in order to accomplish $\T_{MAP}$. Finally, $\A=\bigcup_{i \in \AG}\A^i$ is the set of planning actions of the agents. $\A^i$ and $\A^j$ of two specialized agents $i$ and $j$ will typically be two disjoint sets since the agents have their own different capabilities; otherwise, $\A^i$ and $\A^j$ may overlap. $\A$ includes two fictitious actions $\alpha_i$ and $\alpha_f$ that do not belong to the action set of any particular agent: $\alpha_i$ represents the initial state of $\T_{MAP}$, i.e., $PRE(\alpha_i)=\emptyset$ and $EFF(\alpha_i)=\I$, while $\alpha_f$ represents the global goals of $\T_{MAP}$, i.e., $PRE(\alpha_f)=\G$, and $EFF(\alpha_f)=\emptyset$.
\end{definition}

\vspace{0,1cm}

As discussed in Example \ref{example_task}, our model considers specialized agents that can be functionally and/or spatially distributed. This specialization defines the local \emph{view} that each agent has of the MAP task. Local views are a typical characteristic of multi-agent systems and other distributed systems. For instance, distributed CSPs use local views, such that agents only receive information about the constraints in which they are involved \cite{Jannach13,Yokoo98}. Next, we define the information of an agent $i$ on a planning task $\T_{MAP}$.

The \emph{view} of an agent $i$ on a MAP task $\T_{MAP}$ is defined as $\T^i_{MAP} =  \langle\Prop^i,\A^i,\I^i,\G\rangle$. $\Prop^i$ is the set of state variables known to agent $i$; $\A^i \subseteq \A$ is the set of its capabilities (planning actions); $\I^i$ is the subset of fluents of the initial state $\I$ that are visible to agent $i$; and $\G$ is the set of global goals of $\T_{MAP}$. Since agents in FMAP are fully cooperative, they are all aware of the global goals of the task. Obviously, because of specialization, a particular agent may not understand the goals as specified in $\G$; defining $\G$ as global goals implies that all agents contribute to the achievement of $\G$, either directly (achieving a $g \in \G$) or indirectly (introducing actions whose effects help other agents achieve $g$).

The state variables of an agent $i$ are determined by the view the agent has on the initial state, $I^i$, the planning actions it can perform, $A^i$, and the set of goals of $\T_{MAP}$. This also affects the domain $D_v$ of a variable $v$. We define $\D_{v}^i \subseteq \D_v$ as the set of values of the variable $v$ that are known to agent $i$.

Consider again the $pos\mhyphen rm$ variable in Example \ref{example_task}. The domain of $pos\mhyphen rm$ contains all the locations in the transportation task, including the factory $f$, the storage facility $sf$, and the trucks; that is, $\D_{pos\mhyphen rm}=\{l1,l2,l3,l4,f,sf,t1,t2\}$. However, agents $ta1$ and $ta2$ have local knowledge about the domain of $pos\mhyphen rm$ because some of the values of such variable refer to objects of $\T_{MAP}$ that are unknown to them. Hence, $ta1$ will manage $\D^{ta1}_{pos\mhyphen rm}=\{l1,l2,sf,t1\}$, while $ta2$ will manage $\D^{ta2}_{pos\mhyphen rm}=\{l3,l4,sf,f,t2\}$.

Agents in FMAP interact with each other by sharing information about their state variables. For each pair of agents $i$ and $j$, the public information they share is defined as $\Prop^{ij} = \Prop^{ji}=\Prop^i\cap\Prop^j$. Additionally, some of the values in the domain of a variable can also be public to both agents. The set of values of a variable $v$ that are public to a pair of agents $i$ and $j$ is defined as $\D^{ij}_v = \D^i_v \cap \D^j_v$.

As Example \ref{example_task} indicates, the $pos\mhyphen rm$ variable is public to agents $ta1$ and $ta2$. The values that are public to both agents are defined as the intersection of the values that are known to each of them, $\D^{ta1\;ta2}_{pos\mhyphen rm} = \{sf\}$. This way, the only public location of $rm$ for agents $ta1$ and $ta2$ is the storage facility $sf$, which is precisely the intersection between the two geographical areas. Hence, if agent $ta1$ places $rm$ in $sf$, it will inform $ta2$ accordingly, and vice versa. This allows agents $ta1$ and $ta2$ to work together while minimizing the information they share with each other.

\vspace{0,1cm}

Our MAP model is a multi-agent refinement planning framework, which is a general method based on the refinement of the set of all possible plans. The internal reasoning of agents in FMAP is configured as a Partial-Order Planning (POP) search procedure. Other local search strategies are applicable, as long as agents build \emph{partial-order plans}. The following concepts and definitions are standard terms from the POP paradigm \cite{Ghallab04}, which have been adapted to state variables. Additionally, definitions also account for the multi-agent nature of the planning task and the local views of the task by the agents.

\vspace{0,1cm}
\begin{definition}
A \textbf{partial-order plan} or partial plan is a tuple $\Pi = \langle \Delta, \Ord, \CL \rangle$. $\Delta = \{ \alpha | \alpha \in \A \}$ is the set of actions in $\Pi$. $\Ord$ is a finite set of ordering constraints ($\prec$) on $\Delta$. $\CL$ is a finite set of causal links of the form $\alpha \stackrel{\langle v,d\rangle}{\rightarrow} \beta$ or $\alpha \stackrel{\langle v,\neg d\rangle}{\rightarrow} \beta$, where $\alpha$ and $\beta$ are actions in $\Delta$. A causal link $\alpha \stackrel{\langle v,d\rangle}{\rightarrow} \beta$ enforces precondition $\langle v, d\rangle \in PRE(\beta)$ through an effect $(v = d) \in EFF(\alpha)$ \cite{Ghallab04}. Similarly, a  causal link $\alpha \stackrel{\langle v,\neg d\rangle}{\rightarrow} \beta$ enforces $\langle v, \neg d\rangle \in PRE(\beta)$ through an effect $(v \not= d) \in EFF(\alpha)$ or $(v = d') \in EFF(\alpha)$, $d' \not= d$.
\end{definition}
\vspace{0,1cm}

An \emph{empty} partial plan is defined as $\Pi_0 = \langle \Delta_0$, $\Ord_0$, $\CL_0 \rangle$, where  $\Ord_0$ and $\CL_0$ are empty sets, and $\Delta_0$ contains only the fictitious initial action $\alpha_i$. A partial plan $\Pi$ for a task $\T_{MAP}$ will always contain $\alpha_i$.

The introduction of new actions in a partial plan may trigger the appearance of \emph{flaws}. There are two types of flaws in a partial plan: preconditions that are not yet solved (or supported) through a causal link, and threats. A \emph{threat} over a causal link $\alpha \stackrel{\langle v,d\rangle}{\rightarrow} \beta$ is caused by an action $\gamma$ that is not ordered w.r.t. $\alpha$ or $\beta$ and might potentially modify the value of $v$ \cite{Ghallab04} ($(v \not= d) \in EFF(\gamma)$ or $(v=d') \in EFF(\gamma)$, $d' \neq d$), making the causal link unsafe. Threats are addressed by introducing either an ordering constraint $\gamma \prec \alpha$ (this is called \emph{demotion} because the causal link is posted after the threatening action) or an ordering $\beta \prec \gamma$  (this is called \emph{promotion} because the causal link is placed before the threatening action) \cite{Ghallab04}.

A \emph{flaw-free} plan is a threat-free partial plan in which the preconditions of all the actions are supported through causal links.

\vspace{0,2cm}
Planning agents in FMAP cooperate to solve MAP tasks by progressively refining an initially empty plan $\Pi$ until a solution is reached. The definition of \emph{refinement plan} is closely related to the internal forward-chaining partial-order planning search performed by the agents. Refinement planning is a technique that is widely used by many planners, specifically in anytime planning, where a first initial solution is progressively refined until the deliberation time expires \cite{Sapena08b}. We define a refinement plan as follows:

\vspace{0,1cm}
\begin{definition}
A \textbf{refinement plan} $\Pi_r = \langle \Delta_r$, $\Ord_r$, $\CL_r \rangle$ over a partial plan $\Pi = \langle \Delta$, $\Ord$, $\CL \rangle$ is a flaw-free partial plan that extends $\Pi$, i.e., $\Delta \subset \Delta_r$, $\Ord \subset \Ord_r$ and $\CL \subset \CL_r$. $\Pi_r$ introduces a new action $\alpha \in \Delta_r$ in $\Pi$, resulting in $\Delta_r = \Delta \cup \alpha$. All the preconditions in $PRE(\alpha)$ are linked to existing actions in $\Pi$ through causal links; i.e., all preconditions are supported: $\forall p \in PRE(\alpha)$, $\exists$ $\beta \stackrel{p}{\rightarrow} \alpha \in \CL_r$, where $\beta \in \Delta$.
\end{definition}
\vspace{0,1cm}

Refinement plans in FMAP include actions that can be executed in parallel by different agents. Some MAP models consider that two parallel or non-sequential actions are mutually consistent if neither of them modifies the value of a state variable that the other relies on or affects \cite{Brenner09}. We also consider that the preconditions of two mutually consistent actions have to be consistent \cite{Boutilier01}. Hence, two non-sequential actions $\alpha \in \A^i$ and $\beta \in \A^j$ are \emph{mutually consistent} if none of the following conditions hold:

\begin{itemize}
\item $\;\exists (v = d) \in EFF(\alpha)$ and $\exists (\langle v, d'\rangle \in PRE(\beta) \vee \langle v, \neg d\rangle \in PRE(\beta))$, where $v \in \Prop^{ij}$, $d \in \D^{ij}_v $, $d' \in \D^{j}_v$ and $d \not= d'$, or vice versa; that is, the effects of $\alpha$ and the preconditions of $\beta$ (or vice versa) do not contradict each other under the specified conditions.
\item $\;\exists (v = d) \in EFF(\alpha)$ and $\exists ((v = d') \in EFF(\beta) \vee (v \not= d) \in EFF(\beta))$, where $v \in \Prop^{ij}$, $d \in \D^{ij}_v$, $d' \in \D^{j}_v$ and $d \not= d'$, or vice versa; that is, the effects of $\alpha$ and the effects of $\beta$ (or vice versa) do not contradict each other under the specified conditions.
\item $\;\exists \langle v, d\rangle \in PRE(\alpha)$ and $\exists (\langle v, d'\rangle \in PRE(\beta) \vee \langle v, \neg d\rangle \in PRE(\beta))$, where $v \in \Prop^{ij}$, $d \in \D^{ij}_v$, $d' \in \D^{j}_v$ and $d \not= d'$, or vice versa; that is, the preconditions of $\alpha$ and the preconditions of $\beta$ (or vice versa) do not contradict each other under the specified conditions.
\end{itemize}

Agents address parallelism by the resolution of threats over the causal links of the plan. Thus, consistency between any two non-sequential actions introduced by different agents is always guaranteed as refinement plans are flaw-free plans.

Finally, a \emph{solution plan} for $\T_{MAP}$ is a refinement plan $\Pi= \langle \Delta$, $\Ord$, $\CL \rangle$ that addresses all the global goals $\G$ of $\T_{MAP}$. A solution plan includes the fictitious final action $\alpha_f$ and ensures that all its preconditions (note that $PRE(\alpha_f) = \G$) are satisfied; that is, $\forall g \in PRE(\alpha_f)$, $\exists$ $\beta \stackrel{g}{\rightarrow} \alpha_f \in \CL$,  $\beta \in \Delta$, which is the necessary condition to guarantee that $\Pi$ solves $T_{MAP}$.

\subsection{Privacy in partial plans}
\label{privacy}

Every time an agent $i$ refines a partial plan by introducing a new action $\alpha \in \A^i$, it communicates the resulting refinement plan to the rest of the agents in $\T_{MAP}$. As stated above, the information that is public to a pair of agents is defined according to the common state variables and domain values. In order to preserve privacy, agent $i$ will only communicate to agent $j$ the fluents in action $\alpha$ whose variables are common to both agents. The information of a refinement plan $\Pi$ that agent $j$ receives from agent $i$ configures its \emph{view} of that plan, $view^j(\Pi)$. More specifically, given two agents $i$ and $j$ and a fluent $\langle v,d\rangle$, where $v \in \Prop^i$ and $d \in \D^i_{v}$ (equivalently for a negative fluent $\langle v,\neg d\rangle$), we distinguish the three following cases: 

\begin{itemize}
	\item \textbf{Public fluent}: if $v \in \Prop^{ij}$ and $d \in \D^{ij}_v$, the fluent $\langle v,d\rangle$ is public to both $i$ and $j$, and thus agent $i$ will send agent $j$ all the causal links, preconditions, and effects regarding $\langle v, d \rangle$.
	\item \textbf{Private fluent to agent $i$}: if $v \not\in \Prop^{ij}$, the fluent $\langle v,d\rangle$ is private to agent $i$ w.r.t. agent $j$, and thus agent $i$ will occlude the preconditions and effects regarding $\langle v, d \rangle$ to agent $j$. Causal links of the form $\alpha \stackrel{\langle v,d\rangle}{\rightarrow} \beta$ will be sent to agent $j$ as ordering constraints $\alpha \prec \beta$.
	\item \textbf{Partially private fluent to agent $i$}: if $v \in \Prop^{ij}$ but $d \not\in \D^{ij}_v$, the fluent $\langle v,d\rangle$ is partially private to agent $i$ w.r.t. agent $j$. Instead of $\langle v,d\rangle$, agent $i$ will send agent $j$ a fluent $\langle v, \perp \rangle$, where $\perp$ is the \emph{undefined value}. Hence, preconditions of the form $\langle v,d\rangle$ will be sent as $\langle v,\perp\rangle$, effects of the form $(v=d)$ will be replaced by $(v=\perp)$, and causal links $\alpha \stackrel{\langle v,d\rangle}{\rightarrow} \beta$ will adopt the form $\alpha \stackrel{\langle v,\perp\rangle}{\rightarrow} \beta$.
\end{itemize}

If an agent $j$ receives a fluent $\langle v, \perp \rangle$, $\perp$ is interpreted as follows: $\forall d \in \D^j_v, \; \langle v, \neg d\rangle$. That is, $\perp$ indicates that $v$ is not assigned any of the values known to agent $j$ ($D^j_v$). This mechanism is used to inform an agent that a resource is no longer available in its influence area. For instance, suppose that agent $ta2$ in Example \ref{example_task} acquires the raw material $rm$ from $sf$ by loading it into its truck $t2$. Agent $ta2$ communicates to $ta1$ that $rm$ is no longer in $sf$, but agent $ta1$ does not know about the truck $t2$. To solve this issue, $ta2$ sends $ta1$ the fluent $\langle pos\mhyphen rm, \perp\rangle$, meaning that the resource $rm$ is no longer available in the geographical area of agent $ta1$. Consequently, $ta1$ is now aware that $rm$ is not located in any of its accessible positions $D^{ta1}_{pos\mhyphen rm} = \{l1,l2,sf,t1\}$. 

\begin{figure}
\centering
\includegraphics[width=8cm]{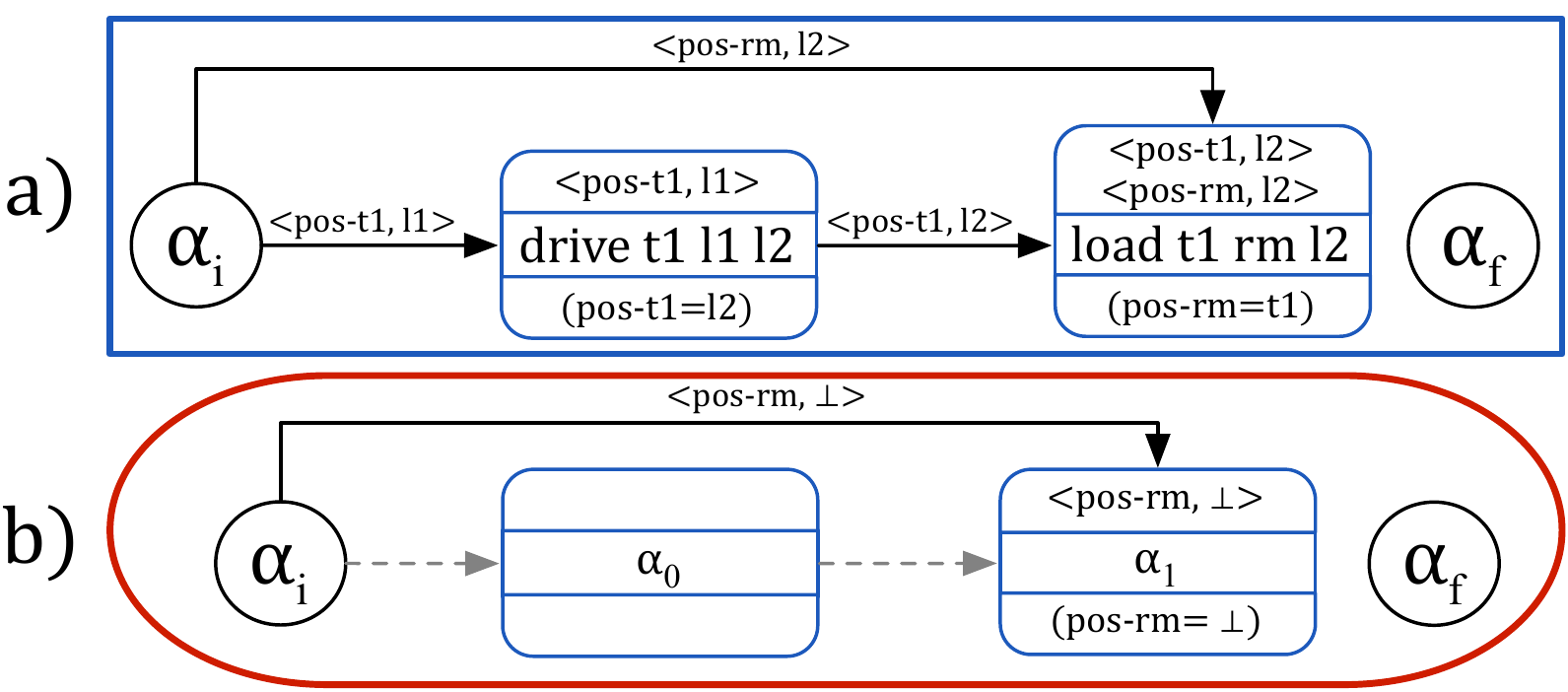}
\caption{A refinement plan $\Pi_r$ as viewed by: a) agent $ta1$ b) agent $ta2$}
\label{view}
\end{figure}

\vspace{0.1cm}
Fig. \ref{view} shows the view that the transport agents $ta1$ and $ta2$ in Example \ref{example_task} have of a simple refinement plan $\Pi_r$. In this plan, agent $ta1$ drives the truck $t1$ from $l1$ to $l2$ and loads $rm$ into $t1$. As shown in Fig. \ref{view}a, $view^{ta1}(\Pi_r)$ contains all the information of both actions in the plan since agent $ta1$ has introduced them. Agent $ta2$, however, does not know about the truck $t1$, and hence the variable $pos\mhyphen t1$, which models the position of $t1$, is private to $ta1$ w.r.t. $ta2$. This way, all the preconditions and effects related to the fluents $\langle pos\mhyphen t1, l1\rangle$ and $\langle pos\mhyphen t1, l2\rangle$ are occluded in $view^{ta2}(\Pi_r)$ (see Fig. \ref{view}b). Additionally, the causal links regarding these two fluents are replaced by ordering constraints in $view^{ta2}(\Pi_r)$. On the other hand, the variable $pos\mhyphen rm$ is public to both agents, but the $load$ action refers to the locations $t1$ and $l2$, which are not in $\D^{ta2}_{pos\mhyphen rm}$. Therefore, fluents $\langle pos\mhyphen rm, l2 \rangle$ and $\langle pos\mhyphen rm, t1\rangle$ are partially private to agent $ta1$ w.r.t. $ta2$. This way, in $view^{ta2}(\Pi_r)$, the precondition $\langle pos\mhyphen rm, l2 \rangle$ and the effect $(pos\mhyphen rm = t1)$ of the $load$ action are replaced by $\langle pos\mhyphen rm, \perp \rangle$ and $(pos\mhyphen rm = \perp)$, respectively. The fluent $\langle pos\mhyphen rm, l2 \rangle$ is also replaced by $\langle pos\mhyphen rm, \perp \rangle$ in the causal link $\alpha_i \stackrel{\langle pos\mhyphen rm,l2\rangle}{\rightarrow} load\;t1\;rm\;l2$.

\subsection{MAP definition language}

There is a large body of work on planning task specification languages. Since planning has been traditionally regarded as a centralized problem, the most popular definition languages, such as the different versions of \emph{PDDL} (the Planning Domain Definition Language\footnote{\url{http://en.wikipedia.org/wiki/Planning_Domain_Definition_Language}}), are designed to model single-agent planning tasks. MAP introduces a set of requirements that are not present in single-agent planning, such as privacy or specialized agents, which motivate the development of specification languages for multi-agent planning. 

There are many different approaches to MAP as described in section \ref{related}. \emph{MA-STRIPS} \cite{Brafman08}, which was designed as a minimalistic extension to \emph{STRIPS} \cite{Fikes71}, is one of the most common MAP languages. It allows defining a set of agents and associating the planning actions they can execute. FMAP presents several advanced features that motivated the definition of our own \emph{PDDL}-based specification language (the language syntax is detailed in \cite{Torreno12KAIS}) rather than using \emph{MA-STRIPS}.

Since the world states in FMAP are modeled through state variables instead of predicates, our MAP language is based on \emph{PDDL3.1} \cite{Kovacs11}, the latest version of \emph{PDDL}. Unlike its predecessors, which model planning tasks through predicates, \emph{PDDL3.1} incorporates state variables that map to a finite domain of objects of the task. 

In a single-agent language, the user specifies the \emph{domain} of the task (planning operators, types of objects, and functions) and the \emph{problem} to be solved (objects of the task, initial state, and goals). In FMAP, we write a domain and a problem file for each agent, which define the typology of the agent, and the agent's local view of the MAP task, respectively. The domain files keep the structure of a regular \emph{PDDL3.1} domain file. The problem files, however, are extended with an additional {\ttfamily :shared-data} section, which specifies the information that an agent can share with each of the other participating agents in the task.

\section{FMAP refinement planning procedure}
\label{algorithms}

FMAP is based on a cooperative refinement planning procedure in which agents jointly explore a multi-agent, plan-space search tree. A multi-agent search tree is one in which the partial plans of the nodes are built with the contributions of one or more agents. 

\begin{figure}
\centering
\includegraphics[width=8.5cm]{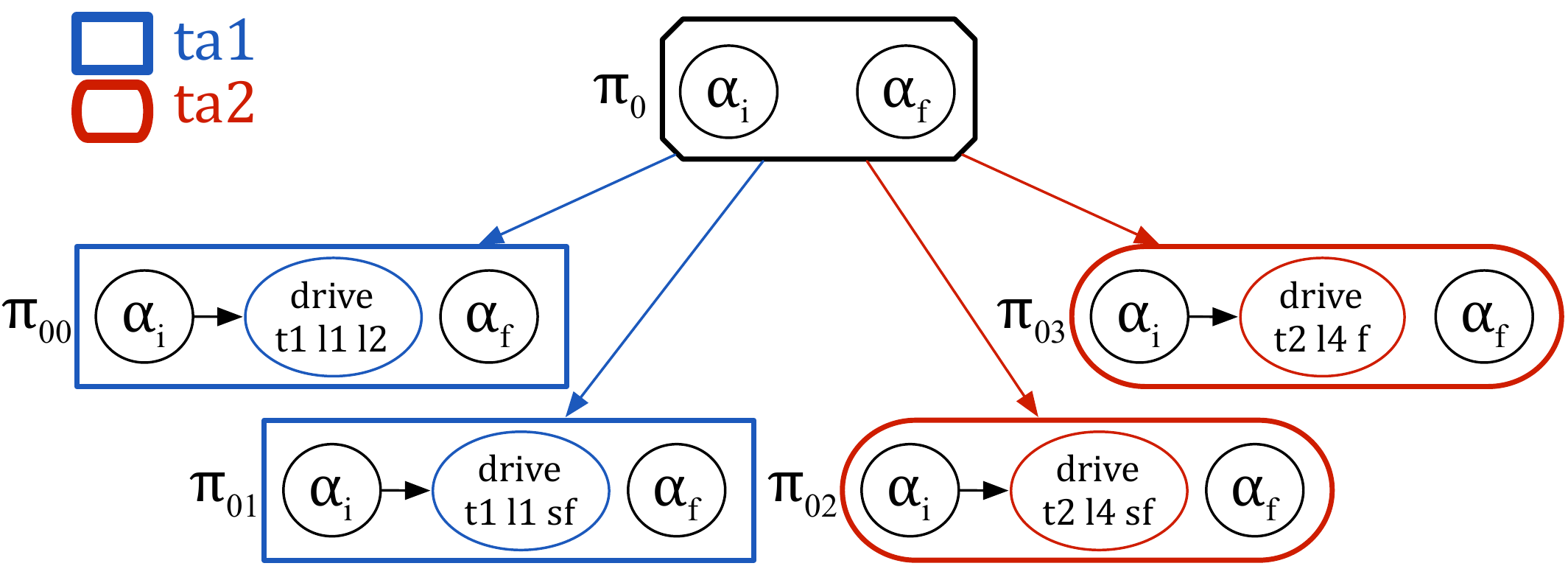}
\caption{FMAP multi-agent search tree example}
\label{tree}
\end{figure}

Fig. \ref{tree} shows the first level of the multi-agent search tree that would be generated for the transportation task of Example \ref{example_task}. At this level, agents $ta1$ and $ta2$ each propose two refinement plans, specifically the plans to move their trucks within their geographical areas. In each of these refinement plans, the agent adds one action and the corresponding orderings and causal links. Agent $f$ does not contribute here with any refinement plan because the initial empty plan $\Pi_0$ does not have the necessary supporting information for $f$ to insert any of its actions. In a subsequent iteration (expansion of the next tree node), agents can in turn create new refinement plans. For instance, if node $\Pi_{00}$ in Fig. \ref{tree} is selected next for expansion, the three agents in the problem ($ta1$, $ta2$, or $f$) will try to create refinement plans over $\Pi_{00}$ by adding one of their actions and supporting it through the necessary causal links and orderings. 
 
Agents keep a copy of the multi-agent search tree, storing the local view they have of each of the plans in the tree nodes. Given a node $\Pi$ in the multi-agent search tree, an agent $i$ maintains $view^i(\Pi)$ in its copy of the tree.

FMAP applies a multi-agent A* search that iteratively explores the multi-agent tree. One iteration of FMAP involves the following: 1) agents select one of the unexplored leaf nodes of the tree for expansion; 2) agents expand the selected plan by generating all the refinement plans over this node; and 3) agents evaluate the resulting successor nodes and communicate the results to the rest of the agents. Instead of using a broadcast control framework, FMAP uses democratic leadership, in which a coordinator role is scheduled among the agents. One of the agents adopts the role of coordinator at each iteration, thus leading the procedure in one iteration (initially, the coordinator role is randomly assigned to one of the participating agents). More specifically, a FMAP iteration is as follows:

\begin{itemize}
	\item \textbf{Base plan selection}: Among all the open nodes (unexplored leaf nodes) of the multi-agent search tree, the coordinator agent selects the most promising plan, $\Pi_b$, as the base plan to refine in the current iteration. $\Pi_b$ is selected according to the evaluation of the open nodes (details on the node evaluation and selection are presented in section \ref{heuristic}). In the initial iteration, the base plan is the empty plan $\Pi_{0}$.
		
	\item \textbf{Refinement plan generation}: Agents expand $\Pi_b$ and generate its successor nodes. A successor node is a refinement plan over $\Pi_b$ that an agent generates individually through its embedded forward-chaining partial-order planner (see subsection \ref{lcf}). 
	
	\item \textbf{Refinement plan evaluation}: Each agent $i$ evaluates its refinement plans $\Pi_r$ by applying a classical A* evaluation function ($f(\Pi_r) = g(view^i(\Pi_r)) + h(view^i(\Pi_r))$). The expression $g(view^i(\Pi_r))$ stands for the number of actions of $\Pi_r$. Since agents view all the actions of the plans (but not necessarily all their preconditions and effects), $g(view^i(\Pi_r))$ is equivalent to $g(\Pi_r)$. $h(view^i(\Pi_r))$ applies our DTG-based heuristic (see subsection \ref{heuristic}) to estimate the cost of reaching a solution plan from $\Pi_r$.
	
	\item \textbf{Refinement plan communication}: Each agent communicates its refinement plans to the rest of the agents. The information that an agent $i$ communicates about its plan $\Pi_r$ to the rest of the agents depends on the level of privacy specified with each of them. Along with the refinement plan $\Pi_r$, agent $i$ communicates the result of the evaluation of $\Pi_r$, $f(\Pi_r)$.
\end{itemize}

Once the iteration is completed, the leadership is handed to another agent, which adopts the coordinator role, and a new iteration starts. The next coordinator agent selects the open node $\Pi$ that minimizes $f(\Pi)$ as the new base plan $\Pi_b$, and then, agents proceed to expand it. This iterative process carries on until $\Pi_b$ becomes a solution plan that supports the final action $\alpha_f$, or when all the open nodes have been visited, in which case, the agents will have explored the complete search space without finding a solution for the MAP task $\T_{MAP}$.

A refinement plan $\Pi$ is evaluated only by the agent that generates it. The agent communicates $\Pi$ along with $f(\Pi)$ to the rest of the agents.  Therefore, the decision on the next base plan is not affected by the agent that plays the coordinator role since all of the agents manage the same $f(\Pi)$ value for every open node $\Pi$.

In the example depicted in Fig. \ref{tree}, agent $ta1$ evaluates its refinement plans, $\Pi_{00}$ and $\Pi_{01}$, and communicates them along with $f(\Pi_{00})$ and $f(\Pi_{01})$ to agents $ta2$ and $f$; likewise, $ta2$ with $ta1$ and $f$. In this first level of the tree, agents $ta1$ and $ta2$ have a complete view of the refinement plans, that they have generated since these plans only contain an action that they themselves introduced. However, when $ta1$ and $ta2$ communicate their plans to each other, they will only send the fluents according to the level of privacy defined between them, as described in subsection \ref{privacy}. This way, $ta1$ will send $view^{ta2}(\Pi_{00})$ and $view^{ta2}(\Pi_{01})$ to agent $ta2$, and $view^{f}(\Pi_{00})$ and $view^{f}(\Pi_{01})$ to agent $f$.

The following subsections analyze the key elements of FMAP, that is, the search algorithm that agents use for the generation of the refinement plans and the heuristic function they use for plan evaluation. We also include a subsection that addresses the completeness and correctness of the algorithm as well as a subsection that describes the limitations of FMAP. 

\subsection{Forward-Chaining Partial-Order Planning}
\label{lcf}

Agents in FMAP use an embedded flexible forward-chaining POP system to generate the refinement plans; this will be referred to as FLEX in the remainder of the paper. Similarly to other approaches, FLEX explores the potential of forward search to support partial-order planning. OPTIC \cite{Benton12}, for instance, combines partial-order structures with information on the \emph{frontier state} of the plan. Informally speaking, the frontier state of the partial plan of a tree node is the resulting state after executing the actions in such a plan. Given a refinement plan $\Pi= \langle \Delta, \Ord, \CL \rangle$, we define its \emph{frontier state} $FS(\Pi)$ as the set of fluents $ \langle v,d\rangle$ achieved by actions $\alpha \in \Delta \;|\; \langle v,d\rangle \in EFF(\alpha)$, such that any action $\alpha' \in \Delta$ that modifies the value of the variable $v$ ($\langle v,d' \rangle \in EFF(\alpha') \;|\; d \neq d'$) is not reachable from $\alpha$ by following the orderings and causal links in $\Pi$.

The only actions that OPTIC adds to a plan are those whose preconditions hold in the frontier state. This behaviour forces OPTIC to some early commitments; however, this does not sacrifice completeness, because search can backtrack. Also, TFPOP \cite{Kvarnstrom11} applies a centralized forward-chaining POP for multiple agents, keeping a sequential execution thread per agent.

\begin{figure}
\centering
\includegraphics[width=8.5cm]{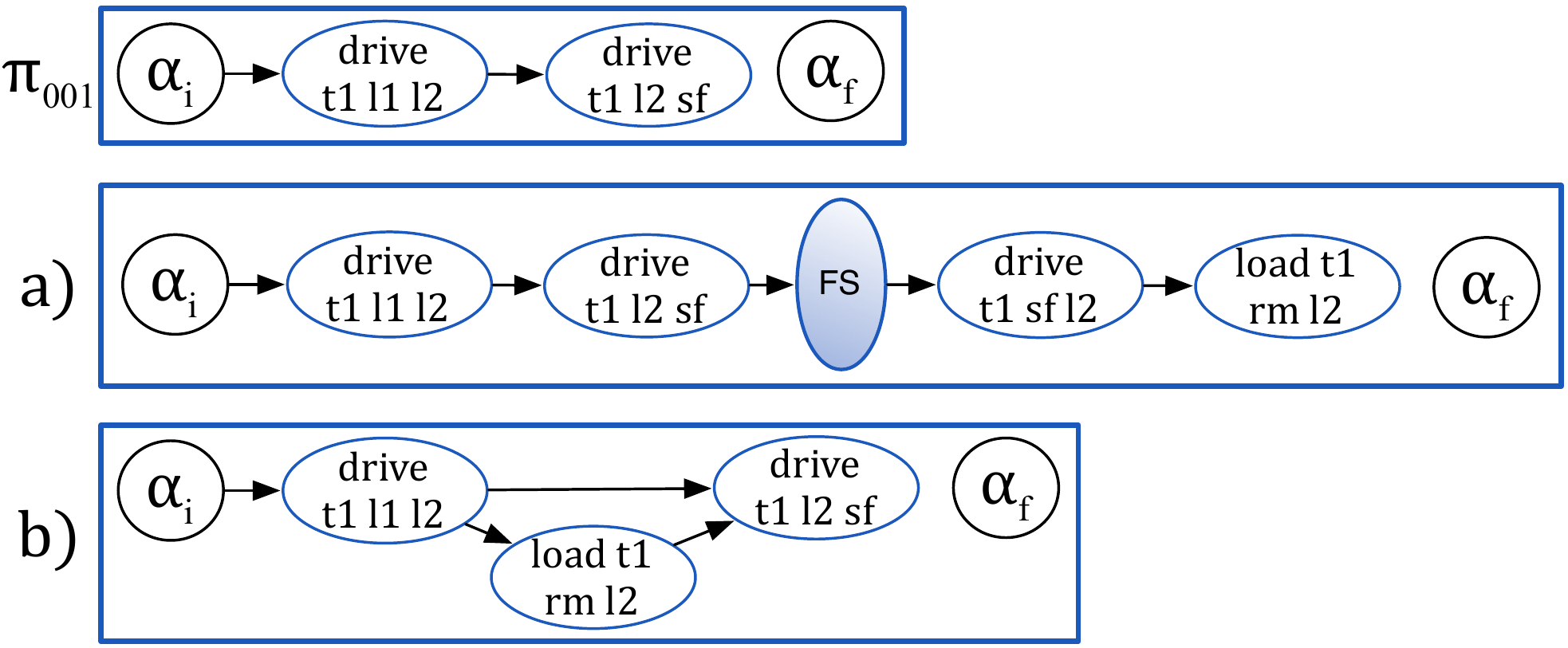}
\caption{Loading $rm$ in plan $\Pi_{001}$: a) inserting actions from a frontier state b) inserting actions using FLEX}
\label{leastCommitment}
\end{figure}

The aforementioned approaches only permit introducing actions that are applicable in the frontier state of the plan. In contrast, FLEX allows inserting actions at any position of the plan without assuming that any action in the plan has already been executed. This is a more flexible approach that is also more compliant with the least-commitment principle that typically guides backward-chaining POP. Fig. \ref{leastCommitment} shows the advantages of our flexible search strategy. Consider the refinement plan $\Pi_{001}$, which is the result of a refinement of agent $ta1$ on plan $\Pi_{00}$ (see Fig. \ref{tree}) after including the action $(drive\;t1\;l1\;sf)$. This is not the best course of action for taking the raw material $rm$ to the factory $f$ as $ta1$ should load $rm$ into $t1$ before moving to $sf$. The frontier state $FS(\Pi_{001})$ reflects the state of the world after executing the plan $\Pi_{001}$, in which the truck $t1$ would be at $sf$. Planners like OPTIC would only introduce actions that are applicable in the frontier state $FS(\Pi_{001})$. In this example, OPTIC would first insert the action $(drive\;t1\;sf\;l2)$ to move the truck $t1$ back to $l2$ in order to be able to apply the action $(load\;t1\;rm\;l2)$ (see Fig. \ref{leastCommitment}a). FLEX, however, is able to introduce actions at any position in the plan, so the $load$ action can be directly placed between both $drive$ actions, thus minimizing the length of the plan (see Fig. \ref{leastCommitment}b).

\begin{algorithm}
\SetInd{0.2em}{0.65em}
\DontPrintSemicolon
\caption{FLEX search algorithm for an agent $i$}
\label{Flex}
	$RP^i \gets \emptyset$ \;
	\If {$potentiallySupportable(\alpha_f, view^i(\Pi_b))$} {
		\Return{$solutionPlans$}
	}
	$CandidateActions \gets \emptyset$ \;
	\ForAll{$\alpha \in \A^i$} {
		\If {$potentiallySupportable(\alpha, view^i(\Pi_b))$} {
			$CandidateActions \gets CandidateActions \cup \alpha$ \;
		}
	}
	\ForAll{$\alpha \in CandidateActions$} {
			$Plans \gets \{view^i(\Pi_b)\}$\;
			\Repeat{$Plans = \emptyset$}{	
				Select and extract $\Pi_s \in Plans$\;
				$Flaws(\Pi_s) \gets unsupportedPrecs(\alpha, \Pi_s) \cup Threats(\Pi_s)$\;
				\If {$Flaws(\Pi_s) = \emptyset$} {
					$RP^i \gets RP^i \cup \Pi_s$\;
				}
				\Else {
					Select and extract $\Phi \in Flaws(\Pi_s)$\;
					$Plans \gets Plans \cup solveFlaw(\Pi_s, \Phi)$\;
				}
		}
	}
	\Return{$RP^i$}
\end{algorithm}

\begin{figure}
\centering
\includegraphics[width=8.5cm]{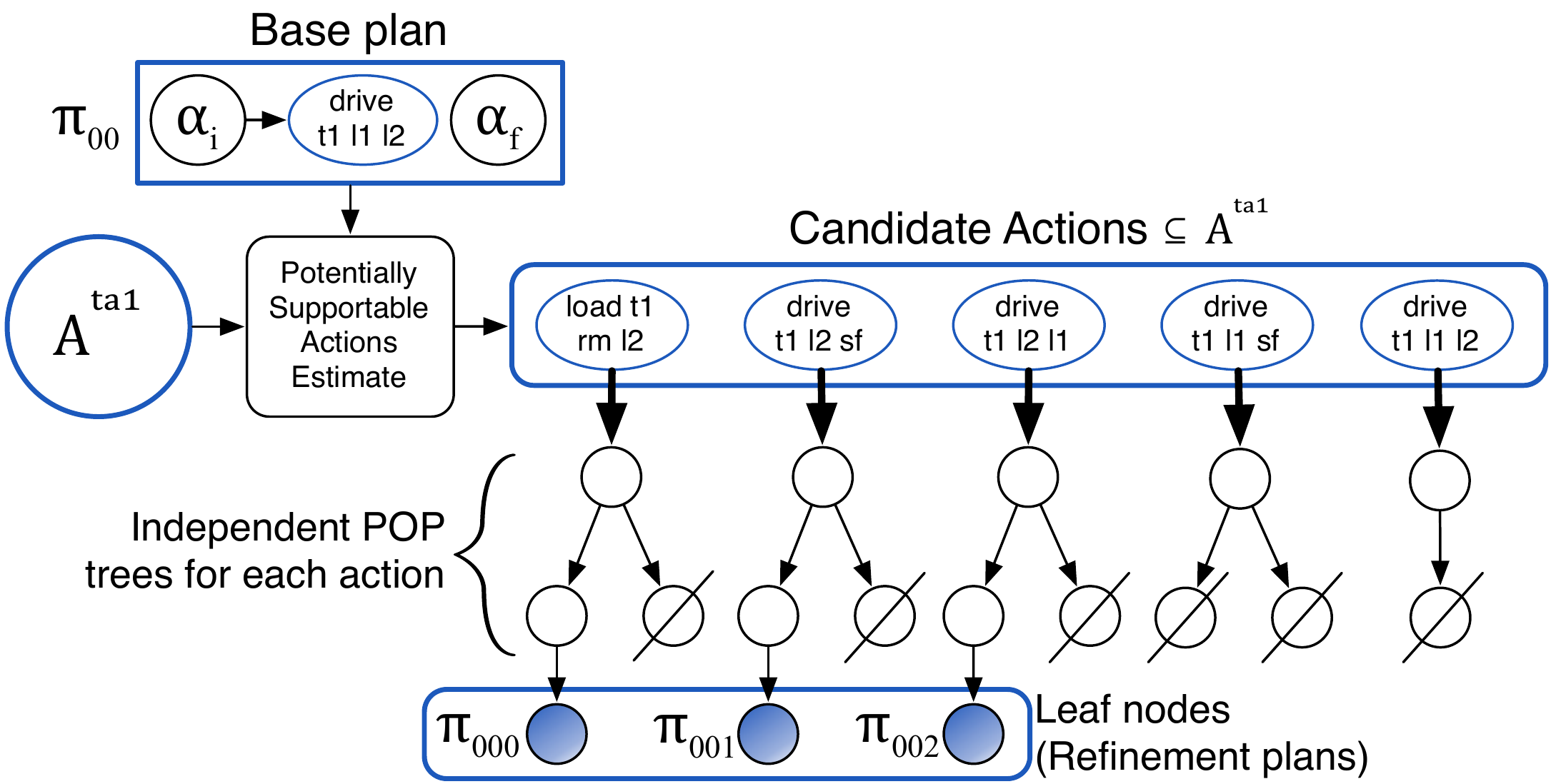}
\caption{FLEX algorithm as applied by agent $ta1$ over plan $\Pi_{00}$}
\label{figLcf}
\end{figure}

Algorithm \ref{Flex} summarizes the FLEX procedure invoked by an agent $i$ to generate refinement plans, and Fig. \ref{figLcf} shows how agent $ta1$ in Example \ref{example_task} uses the FLEX algorithm to refine plan $\Pi_{00}$ in Fig. \ref{tree}. The first operation of an agent $i$ that executes FLEX is to check whether the fictitious final action $\alpha_f$ is supportable in $\Pi_b$, that is, if a solution plan can be obtained from $\Pi_b$. If so, the agent will generate a set of solution plans that covers all the possible ways to support the preconditions of $\alpha_f$ through causal links. 

If a solution plan is not found, agent $i$ analyzes all its planning actions $\A^i$ and estimates if they are supportable in $\Pi_b$. Given an action $\alpha \in \A^i$, the function $potentiallySupportable(\alpha,\Pi_b)$ checks if $\forall \langle v, d\rangle \in PRE(\alpha)$, $\exists \beta \in \Delta(\Pi_b) \;|\; (v = d) \in EFF(\beta)$, i.e., the agent estimates that $\alpha$ is supportable if for every precondition of $\alpha$ there is a matching effect among the actions of $\Pi_b$. 

Fig. \ref{figLcf} shows an example of potentially supportable actions. Agent $ta1$ evaluates all the actions in $\A^{ta}$ and finds five candidate actions. In $\alpha_i$, the initial state of $\Pi_{00}$, the truck $t1$ is at location $l1$. Consequently, $ta1$ considers  $(drive\;t1\;l1\;sf)$ and $(drive\;t1\;l1\;l2)$ as potential candidate actions for its refinements. Note that action $(drive\;t1\;l1\;l2)$ is already included in plan $\Pi_{00}$. Actions $(drive\;t1\;l2\;sf)$, $(drive\;t1\;l2\;l1)$, and $(load\;t1\;rm\;l2)$ are also classified as candidates since they are applicable after the action $(drive\;t1\;l1\;l2)$, which is already in plan $\Pi_{00}$.

It is possible to introduce an action multiple times in a plan; for instance, a truck may need to travel back and forth between two different locations several times. For this reason, $ta1$ again considers $(drive\;t1\;l1\;l2)$ as a candidate action when refining $\Pi_{00}$, even if this action is already included in $\Pi_{00}$. By estimating potentially supportable actions in any position of the plan, FLEX follows the least commitment principle and does not leave out any potential refinement plan. 

The $potentiallySupportable$ procedure is an estimate because it does not actually check the possible flaws that arise when supporting an action. Hence, an agent analyzes the alternatives that support each candidate action $\alpha$ by generating a POP search tree for that particular action (\emph{repeat} loop in Algorithm \ref{Flex}). All the leaf nodes of the tree (stored in the $Plans$ list in Algorithm \ref{Flex}) are explored, thereby covering all the possible ways to introduce $\alpha$ in $\Pi_b$.

As in backward-chaining POP, FLEX introduces the action $\alpha$ in $\Pi_b$ by supporting its preconditions through causal links and solving the threats that arise during the search. The set of flaw-free plans obtained from this search are stored in $RP^i$ as valid refinement plans of agent $i$ over $\Pi_b$. This procedure is carried out for each candidate action. Completeness is guaranteed since all the possible refinement plans over a given base plan are generated by the agents involved in $\T_{MAP}$.

Fig. \ref{figLcf} shows that, for every candidate action, $ta1$ performs an independent POP search aimed at supporting the action. Actions $(load\;t1\;rm\;l2)$, $(drive\;t1\;l2\;sf)$, and $(drive\;t1\;l2\;l1)$ lead to three different refinement plans over $\Pi_{00}$: $\{\Pi_{000},$ $\Pi_{001},\Pi_{002}\}$. These plans will then be inserted into $ta1$'s copy of the multi-agent search tree. Agent $ta1$ will also send the information of these plans to agents $ta2$ and $f$ according to the level of privacy defined with each one. $ta2$ and $f$ also store the received plans in their copies of the tree.

Candidate action $(drive\;t1\;l1\;sf)$ does not produce valid refinement plans because it causes an unsolvable threat. This is because truck $t1$ cannot simultaneously move to two different locations from $l1$, which causes a conflict between the existing action $(drive\;t1\;l1\;l2) \in \Delta(\Pi_{00})$ and $(drive\;t1\;l1\;sf)$. Similarly, action $(drive\;t1\;l1\;l2)$ does not yield any valid refinements. The resulting plan would have two actions $(drive\;t1\;l1\;l2)$ in parallel, both of which are linked to $\alpha_i$, which causes an unsolvable threat because $t1$ cannot perform two identical $drive$ actions in parallel.

\subsection{Completeness and Soundness}

As explained in the previous section, agents refine the base plan concurrently by analyzing all of the possible ways to support their actions in the base plan. Since this operation is done by every agent and for all their actions, we can conclude FMAP is a complete procedure that explores the whole search space. 

As for soundness, a partial-order plan is sound if it is a flaw-free plan. The FLEX algorithm addresses inconsistencies among actions in a partial plan by detecting and solving threats. 

When an agent $i$ introduces an action $\alpha$ in a base plan $\Pi$, FLEX studies the threats that $\alpha$ causes in the causal links of $\Pi$ and the threats that the actions of $\Pi$ may cause in the causal links that support the preconditions of $\alpha$. In both cases, $i$ is able to detect all threats whatever its view of the plan is, $view^i(\Pi)$. That is, FMAP soundness is guaranteed regardless of the level of privacy defined between agents.

With regard to the threats caused by the effects of a new action, privacy may prevent the agent from viewing some of the causal links of the plan. Suppose that agent $i$ introduces an action $\alpha_t$ with an effect $(v=d')$ in plan $\Pi$. Additionally, there is a causal link in $\Pi$ of the form $cl = \alpha_0 \stackrel{\langle v,d\rangle}{\rightarrow} \alpha_1$ introduced by an agent $j$; as $cl$ is not ordered with respect to $\alpha_t$, this situation generates a threat. According to $view^i(\Pi)$, agent $i$ may find one of the following situations:

\begin{itemize}
	\item If $\langle v, d \rangle$ is \emph{public} to $i$ and $j$, then $cl$ is in $view^i(\Pi)$, and thus the threat between $cl$ and $\alpha_t$ will be correctly detected and solved by promoting or demoting $\alpha_t$.
	\item If $\langle v, d \rangle$ is \emph{private} to $j$ w.r.t. $i$, then $\alpha_t$ cannot contain an effect $(v=d')$ because $v \not \in \Prop^i$. Therefore, the threat described above can never occur in $\Pi$.
	\item If $\langle v, d \rangle$ is \emph{partially private} to $j$ w.r.t. $i$, then $cl = \alpha_0 \stackrel{\langle v,d\rangle}{\rightarrow} \alpha_1$ will be seen as  $cl = \alpha_0 \stackrel{\langle v,\perp\rangle}{\rightarrow} \alpha_1$ in $view^i(\Pi)$. Since $\perp \neq d$, agent $i$ will be able to detect and address the threat between $\alpha_t$ and $cl$.
\end{itemize}

Consequently, an agent can always detect the arising threats when it adds a new action, $\alpha_t$, in the plan. Now, we should study whether the potential threats caused by actions in $\Pi$ on the causal links that support the action $\alpha_t$ are correctly detected by agent $i$.  Suppose that there is a causal link $cl' = \beta \stackrel{\langle v',e\rangle}{\rightarrow} \alpha_t$, and an action $\gamma$ with an effect $(v'=e')$ which is not ordered with respect to $\alpha_t$. Again, agent $i$ may find itself in three different scenarios according to its view of $(v'=e')$:

\begin{itemize}
	\item If $(v' = e')$ is \emph{public} to $i$ and $j$, the threat between $cl'$ and $\gamma$ will be correctly detected by $i$.
	\item If $(v' = e')$ is \emph{private} to $j$ w.r.t. $i$, then none of the variables in $PRE(\alpha_t)$ are related to $v'$ because $v' \not \in \Prop^i$. Thus, this threat will never arise in $\Pi$.
	\item If $(v' = e')$ is \emph{partially private} to $j$ w.r.t. $i$, $(v'=e')$ will be seen as $(v'=\perp)$ in $view^i(\Pi)$. Since $\perp \neq e$, the threat between $\gamma$ and $cl'$ will be correctly detected by agent $i$.
\end{itemize}

Note that privacy does not prevent agents from detecting and solving threats nor does it affect the complexity of the process. If the fluent is public or partially private, the agent that is refining the plan will be able to detect the threat because it either sees the value of the variable or sees $\perp$, and both contradict the value of the variable in the causal link. If the fluent is private, then there is no such threat. This proves that FMAP is sound. 

\subsection{DTG-based Heuristic Function}
\label{heuristic}

The last aspect of FMAP to analyze is how agents evaluate the refinement plans. FMAP guides the search through a domain-independent heuristic function, as most planning systems do \cite{Rosa13}. It uses the information provided by the frontier states to perform the heuristic evaluation of the plans contained in the tree nodes.

\begin{figure}
\centering
\includegraphics[width=6cm]{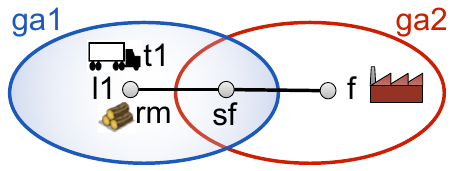}
\vspace{-0.1cm}
\caption{Reduced transport example task}
\label{example2}
\end{figure}

\begin{figure}
\centering
\includegraphics[width=8cm]{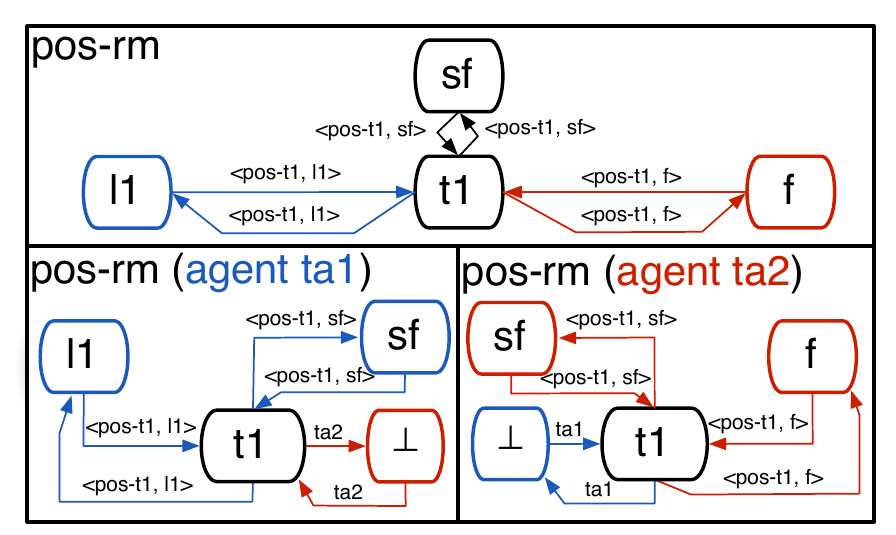}
\caption{Centralized and distributed DTG of the variable $\langle pos\mhyphen rm \rangle$}
\label{dtg}
\end{figure}

According to the definition shown in section \ref{lcf}, the frontier state of a plan $\Pi$, $FS(\Pi)$, can be easily computed as the finite set of fluents that results from executing the actions of the plan $\Pi$ in $\I$, the initial state of $\T_{MAP}$. Since refinement plans are not sequential plans, the actions in $\Delta$ have to be \emph{linearized} in order to compute the frontier state. The linearization of a refinement plan $\Pi$ involves establishing a total order among the actions in $ \Delta$. Given two actions $\alpha \in \Delta$ and $\beta \in \Delta$, if $\alpha \prec \beta \in \Ord$ or $\beta \prec \alpha \in \Ord$, we keep this ordering constraint in the linearized plan. If $\alpha$ and $\beta$ are non-sequential actions, we establish a total ordering among them. Since plans returned by FLEX are free of conflicts, it is irrelevant how non-sequential actions are ordered. 

Frontier states allow us to make use of state-based heuristics such as $h_{FF}$, the relaxed planning graph (RPG) heuristic of FF \cite{Hoffmann01}. However, the distributed approach and the privacy model of FMAP makes the application of $h_{FF}$ inadequate to guide the search. Since none of the agents has knowledge that is complete enough to build an RPG by itself, using $h_{FF}$ to estimate the quality of a refinement plan involves agents building a \emph{distributed RPG} \cite{Feng07}. This is a costly process that requires many communications among agents to coordinate which each other, and it has to be repeated for the evaluation of each refinement plan. Therefore, the predictable high computational cost of the application of $h_{FF}$ led us to discard this choice and opt for designing a heuristic that is based on Domain Transition Graphs (DTGs) \cite{Helmert04}.

A DTG is a directed graph that shows the ways in which a variable can change its value \cite{Helmert04}. Each transition is labeled with the necessary conditions for this to happen; i.e., the preconditions that are common to all the actions that induce the transition. Since DTGs are independent of the state of the plan, recalculations are avoided during the planning process.

Privacy is kept in DTGs through the use of the undefined value $\perp$. This value is represented in a DTG like the rest of the values of the variables, the only difference being that transitions from/to $\perp$ are labeled with the agents that induce them. 

Consider a reduced version of Example \ref{example_task} that is depicted in Fig. \ref{example2}. In this example, both transport agents $ta1$ and $ta2$ can use truck $t1$ within their geographical areas $ga1$ and $ga2$, respectively. Fig. \ref{dtg} shows the DTG of the variable $\langle pos\mhyphen rm \rangle$. In a single-agent task (upper diagram) all the information is available in the DTG. However, in the multi-agent task (bottom diagrams), agent $ta1$ does not know the location of $rm$ if $ta2$ transports it to $f$, while $ta2$ does not know the initial placement of $rm$, since location $l1$ lies outside $ta2$'s geographical area, $ga2$. In order to evaluate the cost of achieving $\langle pos\mhyphen rm, f\rangle$ from the initial state, $ta1$ will first check its DTG, thus obtaining the cost of loading $rm$ in $t1$. As shown in Fig. \ref{dtg}, the transition between values $t1$ and $\perp$ is labeled with agent $ta2$. Therefore, $ta1$ will ask $ta2$ for the cost of the path between values $t1$ and $f$ to complete the calculation. Communications are required to evaluate multi-agent plans, but DTGs are more efficient than RPGs because they remain constant during planning, so agents can minimize the overhead by memorizing paths and distances between values.

For a given plan $\Pi$, our DTG-based heuristic function ($h_{DTG}$ in the following) returns the number of actions of a relaxed plan between the frontier state $FS(\Pi)$ and the set of goals of $\T_{MAP}$, $\G$. $h_{DTG}$ performs a backward search introducing the actions that support the goals in $\G$ into the relaxed plan until all their preconditions are supported. Hence, the underlying principle of $h_{DTG}$ is similar to $h_{FF}$, except for the fact that DTGs are used instead of RPGs to build the relaxed plan.

The $h_{DTG}$ evaluation of a plan $\Pi$ begins by calculating the frontier state $FS(\Pi)$. Next, an iterative procedure is performed to build the relaxed plan. This procedure manages a list of fluents, $openGoals$, initially set to $\G$. The process iteratively extracts a fluent from $openGoals$ and supports it through the introduction of an action in the relaxed plan. The preconditions of such an action are then included in the $openGoals$ list. For each variable $v \in \Prop$, the procedure  manages a list of values, $Values_v$, which is initialized to the value of $v$ in the frontier state $FS(\Pi)$. For each action added to the relaxed plan that has an effect $(v = d')$, $d'$ will be stored in $Values_v$. An iteration of the $h_{DTG}$ evaluation process executes the following stages:

\begin{itemize}
\item \textbf{Open goal selection}: From the $openGoals$ set, the procedure extracts the fluent $\langle v, d_g \rangle \in openG$ that requires the largest number of value transitions to be supported. 

\item \textbf{DTG path computation}: For every value $d_0$ in $Values_v$, this stage calculates the shortest sequence of value transitions in $v$'s DTG from $d_0$ to $d_g$. Each path is computed by applying Dijkstra's algorithm between the nodes $d_0$ and $d_g$ in the DTG associated to variable $v$. The path with the minimum length is stored as $minPath = ((d_0,d_1),(d_1,d_2), \ldots, (d_{g-1}, d_g))$.

\item \textbf{Relaxed plan construction}: For each value transition $(d_i, d_{i+1}) \in minPath$, the minimum-cost action $\alpha_{min}$ that produces such a transition is introduced in the relaxed plan; that is, $\langle v, d_i\rangle \in PRE(\alpha_{min})$ and $(v = d_{i+1}) \in EFF(\alpha_{min})$. The cost of an action is computed as the sum of the minimum number of value transitions required to support its preconditions. The unsupported preconditions of  $\alpha_{min}$ are stored in $openGoals$, so they will be supported in the subsequent iterations. For each effect $(v' = d') \in EFF(\alpha_{min})$,  the value $d'$ is stored in $Values_{v'}$, so $d'$ can be used in the following iterations to support other $openGoals$.
\end{itemize}

The iterative evaluation procedure carries on until all the open goals have been supported, that is, $openGoals = \emptyset$, and $h_{DTG}$ returns the number of actions in the relaxed plan.

\subsection{Limitations of FMAP}
\label{limitations}

In this section, we present some limitations of FMAP that are worth discussing. FMAP builds upon the POP paradigm, so it can handle plans with parallel actions and only enforces an ordering when strictly necessary. FMAP, however, does not yet explicitly manage time constraints nor durative actions. A POP-based planner can easily be extended to incorporate time because the application of the least-commitment principle provides a high degree of execution flexibility. Additionally, POP is independent of the assumption that actions must be instantaneous or have the same duration and allows actions of arbitrary duration and different types of temporal constraints to be defined as long as the conditions under which actions interfere are well defined \cite{Smith00}. In short, POP represents a natural and very appropriate way to include and handle time in a planning framework.

FLEX involves the construction of a POP tree for each potentially supportable action (see Fig. \ref{figLcf}). This procedure is more costly than the operations required by a standard planner to refine a plan. However, the search trees are independent of each other, which makes it possible to implement FLEX by using multiple execution threads. Parallelization improves the performance of FLEX and the ability of FMAP to scale up. Section \ref{results} provides more insight into the FLEX implementation. 

Currently, FMAP is limited to cooperative goals, which means that all the goals are defined as global objectives to all the participating agents (see section \ref{formalization}). Nevertheless, as a future work, we are considering an extension of FMAP to support self-interested agents with local goals.

FMAP is a general procedure aimed at solving any kind of MAP task. In particular, solving tightly-coupled tasks requires a great amount of coordination. Multi-agent coordination in distributed systems where agents must cooperate is always a major issue. This dependency on coordination makes FMAP a communication-reliant approach. Agents not only have to communicate the refinement plans that they build at each iteration, but they also have to communicate during the heuristic evaluation of the refinement plans in order to maintain privacy (see subsection \ref{heuristic}). The usage of a coordinator agent effectively reduces the need for communication. The experimental results will show that FMAP can effectively tackle large problem instances (see section \ref{results}). Nevertheless, reducing communication overhead while keeping the ability to solve any kind of task remains an ongoing research topic that we plan to consider for future developments.

Privacy management is another issue that potentially worsens the performance of FMAP. In section \ref{privacy}, we defined a mechanism to detect and address threats in partial plans, even when agents do not have a complete view of such plans. Privacy does not add extra complexity to FLEX since agents manage the undefined value $\perp$ as any other value in the domain of a variable. It does, however, make the refinement-plan communication stage more complex because, when an agent $i$ sends $view^j(\Pi)$ to an agent $j$, this implies that $i$ must previously adapt the information of $\Pi$ according to the privacy rules defined w.r.t. to $j$.

Privacy also affects the heuristic evaluation of the plans in terms of quality. Since a refinement plan is only evaluated by the agent that generates it and this evaluation is influenced by the agent's view of the plan, the result may not be as accurate as if the agent had had a complete view of the plan. Empirical results, however, will show that, even with these limitations, our heuristic function provides good performance in a wide variety of planning domains (see section \ref{results}).

\section{Experimental results}
\label{results}

In order to assess the performance of FMAP, we ran experimental tests with some of the benchmark problems from the International Planning Competitions\footnote{\url{http://ipc.icaps-conference.org/}} (IPC). More precisely, we adapted the STRIPS problem suites of 10 different domains from the latest IPC editions to a MAP context. The tests compare FMAP with two different state-of-the-art MAP systems: MAPR \cite{Borrajo13} and MAP-POP \cite{Torreno12ECAI}. We excluded Planning First \cite{Nissim10} from the comparison because it is outperformed by MAP-POP \cite{Torreno12ECAI}.

This section is organized as follows: first, we provide some information on the FMAP implementation and experimental setup. Then, we present the features of the tested domains and we analyze the MAP adaptation performed for each domain. Next, we show a comparative analysis between FMAP and the aforementioned planners, MAPR \cite{Borrajo13} and MAP-POP \cite{Torreno12ECAI}. Then, we perform a scalability analysis of FMAP and MAPR. Finally, we summarize and discuss the results obtained by FMAP and how they compare to the other two planners.

\subsection{FMAP implementation and experimental setup}

Most multi-agent applications nowadays make use of middleware multi-agent platforms that provide them with the communication services required by the agents \cite{Ozturk10}. The entire code of FMAP is implemented in Java and builds upon the Magentix2 platform\footnote{\url{http://www.gti-ia.upv.es/sma/tools/magentix2}} \cite{Such12}. Magentix2 provides a set of libraries to define the agents' behavior, along with the communication resources required by the agents. Magentix2 agents communicate by means of the FIPA Agent Communication Language \cite{Obrien98}. Messaging is carried out through the Apache QPid broker\footnote{\url{http://qpid.apache.org/}}, which is a critical component for FMAP agents.

FMAP is optimized to take full advantage of the CPU execution threads. The FLEX procedure, which generates refinement plans over a given base plan, develops a POP search tree for each potentially supportable action of the agent's domain. As the POP trees are completely independent from each other, the processes for building the trees run in parallel for each agent.

Agents synchronize their activities at the end of the refinement plan generation stage. Consequently, FMAP assigns the same number of execution threads to each agent so that they all spend a similar amount of time to complete the FLEX procedure (note that if we allocate extra threads to a subset of the agents, they would still have to wait for the slowest agent to synchronize). FLEX builds as many POP search trees in parallel as execution threads agents have been allocated. The $h_{DTG}$ heuristic is implemented in a similar way. An agent can simultaneously evaluate as many plans as execution threads it has been allocated.

All the experimental tests were performed on a single machine with a quad-core Intel Core i7 processor and 8 GB RAM (1.5 GB RAM available for the Java VM). The CPU used in the experimentation has eight available execution threads, which are distributed as follows: in tasks that involve two agents, FMAP allocates four execution threads per agent; in tasks with three or four agents, each agent has two available execution threads; finally, in tasks involving five or more agents, each agent has a single execution thread at its disposal. For instance, the three agents in Example \ref{example_task} would get two different execution threads in this particular machine. Hence, in the FLEX example depicted in Fig. \ref{figLcf}, agent $ta1$ would be able to study two candidate actions simultaneously, thus reducing the execution time of the overall procedure.

\begin{center}
\begin{table*}
\centering
{\footnotesize
\begin{tabular}{|l||c|c|c|c|c|c|}
	\hhline{-------}
    \multicolumn{1}{|l||}{\multirow{3}{*}{Domain}} &
    \multicolumn{1}{|c|}{\multirow{3}{*}{Typology}} &
    \multicolumn{1}{|c|}{\multirow{3}{*}{IPC}} &
	\multicolumn{1}{|c|}{\multirow{3}{*}{Agents}} &
    \multicolumn{1}{c|}{\multirow{3}{*}{Cooperative goals}} &
    \multicolumn{2}{c|}{Applicability} \\ \cline{6-7}

    \multicolumn{1}{|c||}{} &
    \multicolumn{1}{|c|}{} &
    \multicolumn{1}{|c|}{} &
	\multicolumn{1}{|c|}{} &
    \multicolumn{1}{c|}{} &
    \multicolumn{1}{c|}{\multirow{2}{*}{MAPR}} &
    \multicolumn{1}{c|}{FMAP} \\

    \multicolumn{1}{|c||}{} &
    \multicolumn{1}{|c|}{} &  
    \multicolumn{1}{|c|}{} &
	\multicolumn{1}{|c|}{} &
    \multicolumn{1}{c|}{} &
    \multicolumn{1}{c|}{} &
	\multicolumn{1}{c|}{MAP-POP}\\ \hhline{=======}

	Blocksworld    & Loosely-coupled	& '98 & robot                       		& No  & \cmark & \cmark \\ \cline{1-7}
	Driverlog 	 & Loosely-coupled 	& '02 & driver                      		& No  & \cmark & \cmark \\ \cline{1-7}
	Rovers 	 & Loosely-coupled	& '06 & rover                       		& No  & \cmark & \cmark \\ \cline{1-7}
	Satellite  	 & Loosely-coupled	& '04 & satellite                   		& No  & \cmark & \cmark \\ \cline{1-7}
	Zenotravel 	 & Loosely-coupled 	& '02 & aircraft                    		& No  & \cmark & \cmark \\ \hhline{=======}
	Depots 	 & Tightly-coupled  	& '02 & depot/truck                 		& Yes & \xmark & \cmark \\ \cline{1-7}
	Elevators 	 & Tightly-coupled  	& '11 & fast-elevator/slow-elevator 	& Yes & \xmark & \cmark \\  \cline{1-7}
	Logistics 	 & Tightly-coupled  	& '00 & airplane/truck              		& Yes & \xmark & \cmark \\ \cline{1-7}
	Openstacks 	 & Tightly-coupled  	& '11 & manager/manufacturer    	& Yes & \xmark & \cmark \\  \cline{1-7}
	Woodworking & Tightly-coupled 	& '11 & machine 				& Yes & \xmark & \cmark \\  \cline{1-7}
	\end{tabular}}
\vspace{0.1cm}
\caption{Features of the MAP domains}
\label{features}
\end{table*}
\end{center}

\subsection{Planning domain taxonomy}

The benchmark used for the experiments includes 10 different domains of the IPCs that are suitable for a multi-agent adaptation. The IPC benchmarks come from (potential) real-world applications of planning, and they have become the de facto mechanism for assessing the performance of single-agent planning systems. The \emph{elevators} domain, for instance, is inspired by a real problem of Schindler Lifts Ltd. \cite{Koehler02}; the \emph{satellite} domain is motivated by a NASA space application \cite{Long03}; the \emph{rovers} domain deals with the decision of daily planning activities of Mars rovers \cite{Bresina02}; and the \emph{openstacks} domain is based on the \emph{minimum maximum simultaneous open stacks} combinatorial optimization problem. Hence, all the domains from the IPCs resemble practical scenarios and they are modeled to keep, as much as possible, both their structure and complexity. In MAP, there is not a standardized collection of planning domains available. Instead, MAP approaches adapt some well-known IPC domains to a multi-agent context, namely the \emph{satellite}, \emph{rovers}, and \emph{logistics} domains \cite{Borrajo13,Nissim10,Torreno12ECAI}.

Converting planning domains into a multi-agent version is not always possible due to the domain characteristics. While some IPC domains have a straightforward multi-agent decomposition, others are inherently single-agent. We developed a domain-dependent tool to automatically translate the original STRIPS tasks into our \emph{PDDL}-based MAP language.

The columns in Table \ref{features} describe the main features of the 10 MAP domains that are included in the benchmark. \emph{Typology} indicates whether the MAP tasks of the domain are loosely-coupled or tightly-coupled. \emph{IPC} shows the last edition of the IPC in which the domain was included. \emph{Agents} indicates the types of object used to define the agents. \emph{Cooperative goals} indicates the presence or absence of these goals in the tasks of each domain. Finally, \emph{Applicability} shows the MAP systems that are capable of coping with each domain.

In order to come up with a well-balanced benchmark, we put the emphasis on the presence (or absence) of specialized agents and cooperative goals. Besides the adaptation to a multi-agent context, the 10 selected domains are a good representative sample of loosely-coupled domains with non-specialized agents and tightly-coupled domains with cooperative goals.

Privacy in each domain is defined according to the nature of the problem and the type of agents involved, while maintaining a correlation and identification with the objects in a real-world problem. 

\subsubsection{Loosely-coupled domains}

The five loosely-coupled domains presented in Table \ref{features} are: \emph{Blocksworld}, \emph{Driverlog}, \emph{Rovers}, \emph{Satellite}, and \emph{Zenotravel}. The prime characteristic of these domains is that agents have the same planning capabilities (operators) such that each task goal can be individually solved by a single agent. That is, tasks can be addressed without cooperation among agents. Next, we provide some insight into the features of these domains and the MAP adaptations. 

\emph{Satellite} \cite{Long03}. This domain offers a straightforward multi-agent decomposition \cite{Nissim10,Torreno12ECAI}. The MAP domain features one agent per satellite. The resulting MAP tasks are almost decoupled as each satellite can attain a subset of the task goals (even all the goals in some cases) without interacting with any other agent. The number of agents in the tasks of this domain vary from 1 to 12. The location, orientation, and instruments of a satellite are private to the agent, only the information on the images taken by the satellites is defined as public.

\emph{Rovers} \cite{Long03}. Like the \emph{Satellite} domain, \emph{Rovers} also offers a straightforward decomposition \cite{Nissim10,Torreno12ECAI}. The MAP domain features one agent per rover. Rovers collect samples of soil and rock and hardly interact with each other except when a soil or rock sample is collected by an agent, and so it is no longer available to the rest of the agents. The number of agents ranges from 1 to 8 rovers per task. As in the \emph{Satellite} domain, only the information related to the collected samples is defined as public.

\emph{Blocksworld}. The MAP version of this domain introduces a set of robot agents (four agents per task), each having an arm to arrange blocks. Unlike the original domain, the MAP version of \emph{Blocksworld} allows handling more than one block at a time. All the information in this domain is considered to be public.

\emph{Driverlog} \cite{Long03}. In this MAP domain, the agents are the drivers of the problem, ranging between 2 and 8 agents per task. Driver agents are in charge of driving the available trucks and delivering the packages to the different destinations. All the information in the domain (status of drivers, trucks, and packages) is publicized by the driver agents.

\emph{Zenotravel} \cite{Long03}. This domain defines one agent per aircraft. The simplest tasks include one agent and the most complex ones up to five agents. Aircraft can directly transport passengers to any city in the task. As in the \emph{Blocksworld} and \emph{Driverlog} domains, all the information concerning the situation of the passengers and the current location of each aircraft is publicly available to all the participating agents.

\subsubsection{Tightly-coupled domains}

We also analyzed five additional domains that feature specialized agents with different planning capabilities: \emph{Depots}, \emph{Elevators}, \emph{Logistics}, \emph{Openstacks} and \emph{Woodworking}. The features of these domains give rise to complex, tightly-coupled tasks that require interactions or commitments \cite{Gunay13} among agents in order to be solved. 

\emph{Depots} \cite{Long03}. This domain includes two different types of specialized agents, depots and trucks, that must cooperate in order to solve most of the goals of the tasks. This domain, which is the most complex one in our MAP benchmark, leads to tightly-coupled MAP tasks with many dependences among agents. \emph{Depots} tasks contain a large number of participating agents, ranging from 5 to 12 agents. Only the location of packages and trucks is defined as public information.  

\emph{Elevators}. Each agent in this domain can be a slow-elevator or a fast-elevator. Operators in the STRIPS domain are basically the same for both types of elevators since the differences between them only affect the action costs. Elevator agents, however, are still specialized because the floors they can access are limited. This leads to tasks that require cooperation to fulfill some of the goals. For instance, an elevator may not be able to take a passenger to a certain floor, so it will stop at an intermediate floor so that the passenger can board another elevator that goes to that floor. Tasks include from 3 to 5 agents. Agents share the information regarding the location of the different passengers. 

\emph{Logistics}. This domain presents two different types of specialized agents: airplanes and trucks. The delivery of some of the packages involves the cooperation of several truck and airplane agents (similarly to the example task introduced in this article). Tasks feature from 3 to 10 different agents. Information regarding the position of the packages is defined as public.

\emph{Openstacks} \cite{Gerevini09}. This MAP domain includes two specialized agents in all of the tasks; the \emph{manager} is in charge of handling the orders, and the \emph{manufacturer} controls the different stacks and manufactures the products. Both agents depend on each other to perform their activities, thus resulting in tightly-coupled MAP tasks with inherently cooperative goals. Most of the information regarding the different orders and products is public since both agents need it to interact with each other. 

\emph{Woodworking}. This domain features four different types of specialized agents (a planer, a saw, a grinder and a varnisher) that represent the machines in a production chain. In most cases, the output of one machine constitutes the input of the following one, so \emph{Woodworking} agents have to cooperate to fulfill the different goals. All the tasks include four agents (a machine of each type). All the information on the status of the different wood pieces is publicized since agents require this information in order to operate.

\begin{center}
\begin{table*}
\centering
{\footnotesize
\begin{tabular}{|l||c||c||c|c|c|r||c|c|c|r|}
	\hhline{-----------}
    \multicolumn{1}{|l||}{\multirow{2}{*}{Domain}} &
	\multicolumn{1}{|c||}{\multirow{2}{*}{Tasks}} &
    \multicolumn{1}{c||}{\multirow{2}{*}{Common}} &
    \multicolumn{4}{c||}{FMAP} &
    \multicolumn{4}{c|}{MAPR}  \\ \cline{4-11}

    \multicolumn{1}{|c||}{} &
    \multicolumn{1}{|c||}{} &
	\multicolumn{1}{|c||}{} &
	\multicolumn{1}{c|}{Solved} &
    \multicolumn{1}{c|}{Actions} &
    \multicolumn{1}{c|}{Makespan} &
    \multicolumn{1}{c||}{Time} &
	\multicolumn{1}{c|}{Solved} &
    \multicolumn{1}{c|}{Actions}&
    \multicolumn{1}{c|}{Makespan} &
    \multicolumn{1}{c|}{Time} \\ \hhline{===========}

	Blocksworld 	& 34 & 19 & 19 & 17,79 & 13,68 & 86,17  & 34 & 1,27x & 1,20x & 0,04x  \\ \cline{1-11}
	Driverlog 	& 20 & 15 & 15 & 24,64 & 13,93 & 42,02  & 18 & 1,19x & 1,53x & 0,06x  \\ \cline{1-11}
	Rovers 	& 20 & 19 & 19 & 32,63 & 14,95 & 53,82  & 20 & 0,97x & 0,85x & 0,05x  \\ \cline{1-11}
	Satellite 	& 20 & 15 & 16 & 27,27 & 16,47 & 177,65 & 18 & 1,14x & 1,03x & 0,03x  \\ \cline{1-11}
	Zenotravel 	& 20 & 18 & 18 & 25,50 & 13,94 & 180,62 & 20 & 1,24x & 1,32x & 0,02x  \\ \cline{1-11}
	\end{tabular}}
\vspace{0.1cm}
\caption{Comparison between FMAP and MAPR}
\label{ResMAPR}
\end{table*}
\end{center}

\vspace{-0.7cm}

\subsection{FMAP vs. MAPR comparison}

This subsection compares the experimental results of FMAP and MAPR \cite{Borrajo13}. MAPR is implemented in Lisp and uses LAMA \cite{Richter10} as the underlying planning system, without using a middleware platform for multi-agent systems. Each experiment is limited to 30 minutes.

Table \ref{ResMAPR} shows the comparative results for FMAP and MAPR. The \emph{Solved} columns refer to the number of tasks solved by each approach. The average number of actions, makespan (plan duration), and search time consider only the tasks solved by both FMAP and MAPR (the \emph{Common} column shows the number of tasks solved by both planners). Actions, makespan, and time values in MAPR are relative to the results obtained with FMAP. The values $n$x in Table \ref{Res} indicate "$n$ times as much as the FMAP result". Therefore, an \emph{Actions} or \emph{Makespan} value that is higher than 1x is a better result for FMAP and a value lower than 1x is a worse result for FMAP. However, a \emph{Time} value higher than 1x indicates a better result for FMAP.

Of the most recent MAP systems, MAPR is one that offers excellent performance in comparison to other state-of-the-art MAP approaches \cite{Borrajo13}. However, as reflected in Table \ref{features}, MAPR is only compatible with the loosely-coupled domains in the benchmark. This limitation is due to the planning approach of MAPR. Specifically, MAPR applies a goal allocation procedure, decomposing the MAP task into subtasks and giving each agent a subset of the task goals to solve. Each agent subtask is solved with the single-agent planner LAMA \cite{Richter10} such that the resulting subplans are progressively combined into a global solution. This makes MAPR an incomplete planning approach that is limited to loosely-coupled tasks without cooperative goals. That is, MAPR is built under the assumption that each goal must be addressed by at least one of the agents in isolation \cite{Borrajo13}.

Whereas the communication overhead is relatively high in FMAP (to a large extent, this is due to the use of the Magentix MAS platform), agents in MAPR do not need to communicate during the plan construction because each agent addresses its allocated subgoals by itself. This setup has a rather positive impact on the execution times and the number of problems solved (coverage). As expected, Table \ref{ResMAPR} shows that execution times in MAPR are much lower than FMAP. With respect to coverage, MAPR solves 110 out of 114 loosely-coupled tasks (roughly 96\% of the tasks), while FMAP solves 87 of such tasks (76\%).

However, in most domains, FMAP comes up with better quality plans than MAPR, taking into account the number of actions as well as the makespan. MAPR is limited by the order in which agents solve their subtasks. The first agent that computes a subplan cannot take advantage of the potential synergies that may arise from other agents' actions; the second agent has only the information of the first agent's subplan, and so on. Additionally, the allocation of goals to each agent may lead to poorly balanced plans. Although FMAP is a more time-consuming approach, it avoids these limitations because agents work together to build the plan action by action. Thus, FMAP provides agents with a global view of the plan at each point of the construction process, while agents in MAPR keep a local view of the plan at hand.

The \emph{Driverlog} domain, while being loosely-coupled, offers many possible synergies between agents. For instance, a driver agent can use a truck to travel to its destination and load a package on its way, while another agent may take over the truck and complete the delivery. If the first agent acted in isolation, it would deliver the package and then go back to its destination, which would result in a worse plan. Robot agents in the \emph{Blocksworld} domain can also cooperate to improve the quality of the plans: for instance, a robot can pick up a block so that another robot can retrieve the block below. Goal balance is also a key aspect in \emph{Zenotravel} since aircraft agents have limited autonomy. If an aircraft solves too many goals it may be forced to refuel thereby worsening the plan quality.

\begin{figure*}
\centering
\includegraphics[width=15cm]{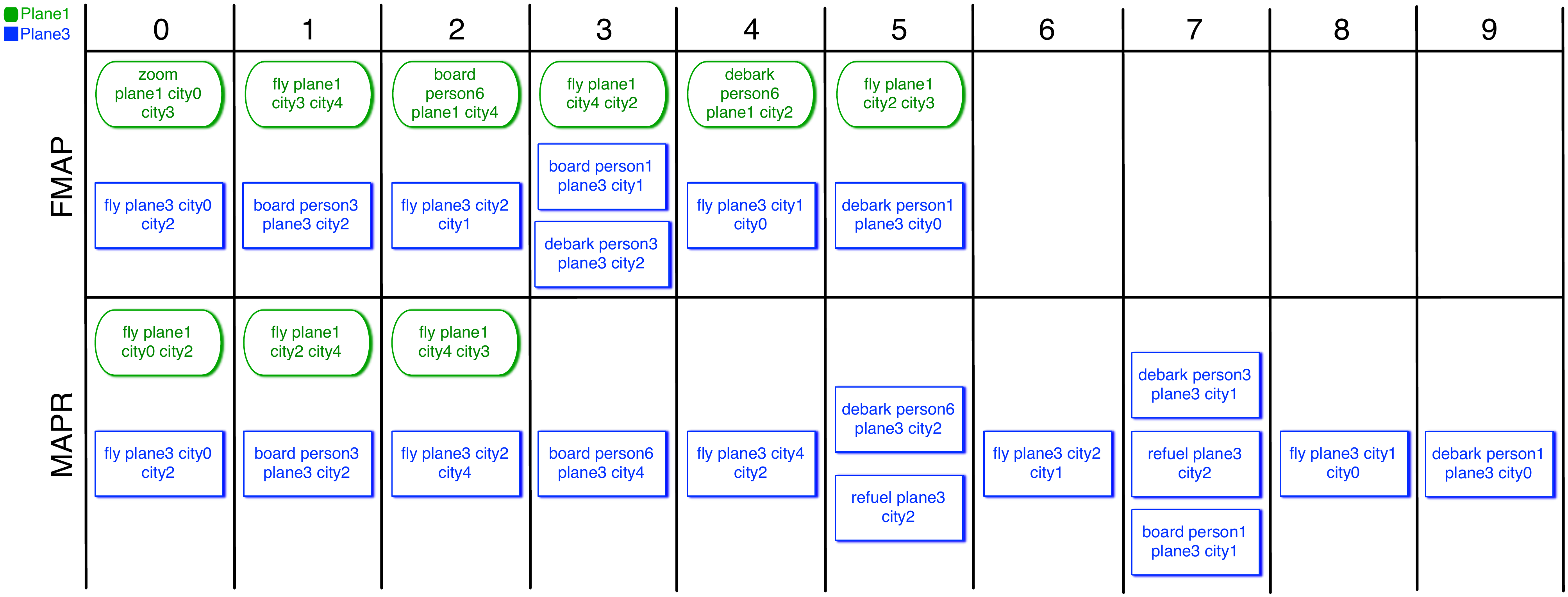}
\caption{\emph{Zenotravel} task 8 solution plan as obtained by FMAP (upper plan) and MAPR (lower plan)}
\label{plan}
\end{figure*}

Fig. \ref{plan} illustrates the MAPR limitations by showing the solution plans obtained by both approaches for task 8 of the \emph{Zenotravel} domain. The goals of this task involve transporting three different people and flying {\ttfamily plane1} to {\ttfamily city3}. The first three goals are achievable by all the plane agents, but the last goal can only be completed by agent {\ttfamily plane1}.

MAPR starts with agent {\ttfamily plane3}, which solves all of the goals that it can. Then, {\ttfamily plane1} receives the subplan and completes it by solving the remaining goal. The resulting joint plan is far from the optimal solution. Agent {\ttfamily plane3} requires 10 time units to solve its subplan because it transports all of the passengers. The high number of {\ttfamily fly} actions forces the agent to introduce additional actions to refuel its tank. On the other hand, agent {\ttfamily plane1} flies directly to its destination without transporting any passengers.

In contrast, agents in FMAP progressively build the solution plan together without using an a-priori goal allocation, which allows them to obtain much better quality plans, taking advantage of synergies  between actions of different agents and effectively balancing the workload among agents. Fig. \ref{plan} shows that, in FMAP, agent {\ttfamily plane1} transports {\ttfamily person6} to its destination, thus simplifying the activities of {\ttfamily plane3}, which avoids refueling. The resulting plan is a much shorter and better balanced solution than the MAPR plan (only 6 time steps versus 10 time steps in MAPR) and it requires fewer actions (13 actions versus 16 in MAPR).

Table \ref{ResMAPR} shows that FMAP noticeably improves plan quality except in the most decoupled domains, namely \emph{Rovers} and \emph{Satellite} (in the latter, FMAP results are slightly better than MAPR results). In these domains, synergies among agents are minimal or even nonexistent. Consequently, MAPR is not penalized by its search scheme, obtaining plans of similar quality to FMAP. 

\begin{center}
\begin{table*}
\centering
{\footnotesize
\begin{tabular}{|l||c||c||c|c|c|r||c|c|c|r|}
	\hhline{-----------}
    \multicolumn{1}{|l||}{\multirow{2}{*}{Domain}} &
    \multicolumn{1}{|c||}{\multirow{2}{*}{Tasks}} &
    \multicolumn{1}{|c||}{\multirow{2}{*}{Common}} &
    \multicolumn{4}{c||}{FMAP} &
    \multicolumn{4}{c|}{MAP-POP} \\ \cline{4-11}

    \multicolumn{1}{|c||}{} &
    \multicolumn{1}{|c||}{} &
    \multicolumn{1}{|c||}{} &
    \multicolumn{1}{c|}{Solved} &
    \multicolumn{1}{c|}{Actions} &
    \multicolumn{1}{c|}{Makespan} &
    \multicolumn{1}{c||}{Time} &
    \multicolumn{1}{c|}{Solved} &
    \multicolumn{1}{c|}{Actions} &
    \multicolumn{1}{c|}{Makespan} &
    \multicolumn{1}{c|}{Time} \\ \hhline{===========}

	Blocksworld    & 34 & 6  & 19 & 9,20 & 7,80 & 7,57 & 6 & 0,91x & 0,74x & 21,49x \\ \cline{1-11}
	Driverlog 	 & 20 & 2 & 15 & 9,50 & 7,00 & 0,66  & 2 & 1,11x & 1,00x & 949,39x \\ \cline{1-11}
	Rovers    	 & 20 & 6 & 19 & 32,63 & 14,95 & 53,82  & 6 & 1,01x & 1,04x & 29,27x \\ \cline{1-11}
	Satellite 	 & 20 & 7 & 16 & 17,14 & 12,57 & 16,00 & 7 & 1,03x & 0,89x & 0,37x \\ \cline{1-11} 
	Zenotravel     & 20 & 3 & 18 & 7,67 & 4,33 & 1,25 & 3 & 1,00x & 1,00x & 87,54x \\ \hhline{===========}
	Depots 	 & 20 & 1   & 6  & 14,00 & 9,00 & 10,56 & 1 & 0,86x & 1,00x & 2,77x \\ \cline{1-11}
	Elevators 	 & 30 & 22 & 30 & 21,32 & 11,36 & 14,60  & 22 & 1,04x & 1,37x & 14,23x \\  \cline{1-11}
	Logistics 	 & 20 & 7 & 10 & 32,29 & 12,71 & 18,26 & 7 & 0,97x & 0,91x & 5,89x \\ \cline{1-11}
	Openstacks 	 & 30 & 0 & 23 & 53,13 & 41,78 & 268,62 & 0  & - & - & - \\ \cline{1-11}
	Woodworking & 30 & 0 & 22 & 16,50 & 4,45 & 100,88 & 0 & -  & - & - \\ \cline{1-11}
	\end{tabular}}
\vspace{0.1cm}
\caption{Comparison between FMAP and MAP-POP}
\label{Res}
\end{table*}
\end{center}

\vspace{-0.7cm}

\subsection{FMAP vs. MAP-POP comparison}
\label{comparison}

We compared FMAP with another recent MAP system, MAP-POP \cite{Torreno12ECAI}. Like FMAP, MAP-POP agents explore the space of multi-agent plans jointly. This set-up allows MAP-POP to overcome some of the limitations of MAPR since it is able to tackle tightly-coupled tasks with cooperative goals. However, MAP-POP has two major disadvantages. Much like MAPR, MAP-POP is an incomplete approach because it implicitly bounds the search tree by limiting its branching factor. This may prevent agents from generating potential solution plans \cite{Torreno12ECAI}. Additionally, MAP-POP is based on backward-chaining POP technologies, thus relying on heuristics that offer a rather poor performance in most MAP domains.

Table \ref{Res} shows the comparison between FMAP and MAP-POP. As in Table \ref{ResMAPR}, the average results consider only the tasks solved by both approaches (the FMAP results for \emph{Openstacks} and \emph{Woodworking} domains include all the tasks solved by this approach because MAP-POP does not solve any of the tasks). The figures in \emph{FMAP} show the results obtained using FMAP for the common problems; \emph{MAP-POP} values are relative to the results of FMAP.

In general, FMAP results are better than MAP-POP results in almost every aspect. In terms of coverage, FMAP clearly outperforms MAP-POP, solving 178 out of 244 tasks (roughly 73\% of the tasks in the benchmark), while MAP-POP solves only 54 tasks (22\%). Overall, in MAP-POP there are problems with some of the most complex tightly-coupled domains (specifically, \emph{Depots}, \emph{Openstacks}, and \emph{Woodworking}), but it performs well in the \emph{Elevators} domain. With respect to the loosely-coupled domains, MAP-POP attains only the simplest tasks, solving from three to seven tasks per domain.

It is difficult to compare the results related to plan quality due to the low coverage of MAP-POP. Focusing on the domains in which MAP-POP solves a significant number of tasks, we observe that MAP-POP obtains slightly better solution plans than FMAP in \emph{Blocksworld} and \emph{Satellite}. FMAP, however, outperforms MAP-POP in \emph{Elevators}, the domain in which both approaches solve the largest number of tasks. 

Finally, the results show that FMAP is much faster than MAP-POP, from 5 times faster in \emph{Logistics} to even 1000 times faster in the \emph{Driverlog} domain. MAP-POP only obtains faster times than FMAP in the seven \emph{Satellite} tasks.

\subsection{Scalability analysis}

\begin{figure}
\centering
\includegraphics[width=6cm]{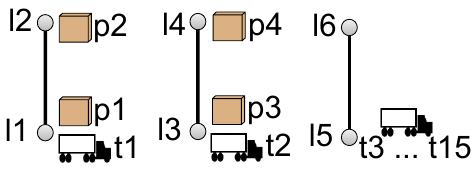}
\caption{\emph{Logistics}-like scalability task}
\label{scalog}
\end{figure}

We prepared two additional experiments to analyze the ability of FMAP and MAPR to scale up. The first test analyzes how both planners scale up when the number of agents of a task is increased, keeping the rest of the parameters unchanged. More specifically, we designed a loosely-coupled \emph{logistics}-like transportation task, which is shown in Fig. \ref{scalog}. The basic task includes two different trucks, $t1$ and $t2$. Truck $t1$ moves between locations $l1$ and $l2$, and truck $t2$ moves between locations $l3$ and $l4$; there is no connection between $t1$'s and $t2$'s locations. The trucks have to transport a total of four packages, $p1\ldots p4$, as shown in Fig \ref{scalog}. In order to ensure that MAPR is able to solve the task, both $t1$ and $t2$ can solve two of the four problem goals by themselves: $t1$ will deliver $p1$ and $p2$, while $t2$ will transport $p3$ and $p4$. Therefore, cooperation is not required in this task, as opposed to the IPC \emph{logistics} domain.

We defined and ran 14 different tests for this basic task. In each test, the number of agents in the task is increased by one, ranging from 2 to 15 truck agents. The problems are modeled so that the extra truck agents, $t3\ldots t15$, are placed in a separate location $l5$, from which there is no access to the locations that $t1$ and $t2$ can move through. Therefore, the additional agents included in each task are unable to solve any of the task goals. However, they do propose refinement plans in FMAP (more precisely, they introduce an action to move to $l6$, as shown in Fig. \ref{scalog}), increasing the complexity of the task in terms of both the number of messages exchanged and the branching factor of the FMAP search tree.

\begin{figure}
\centering
\includegraphics[width=8.5cm]{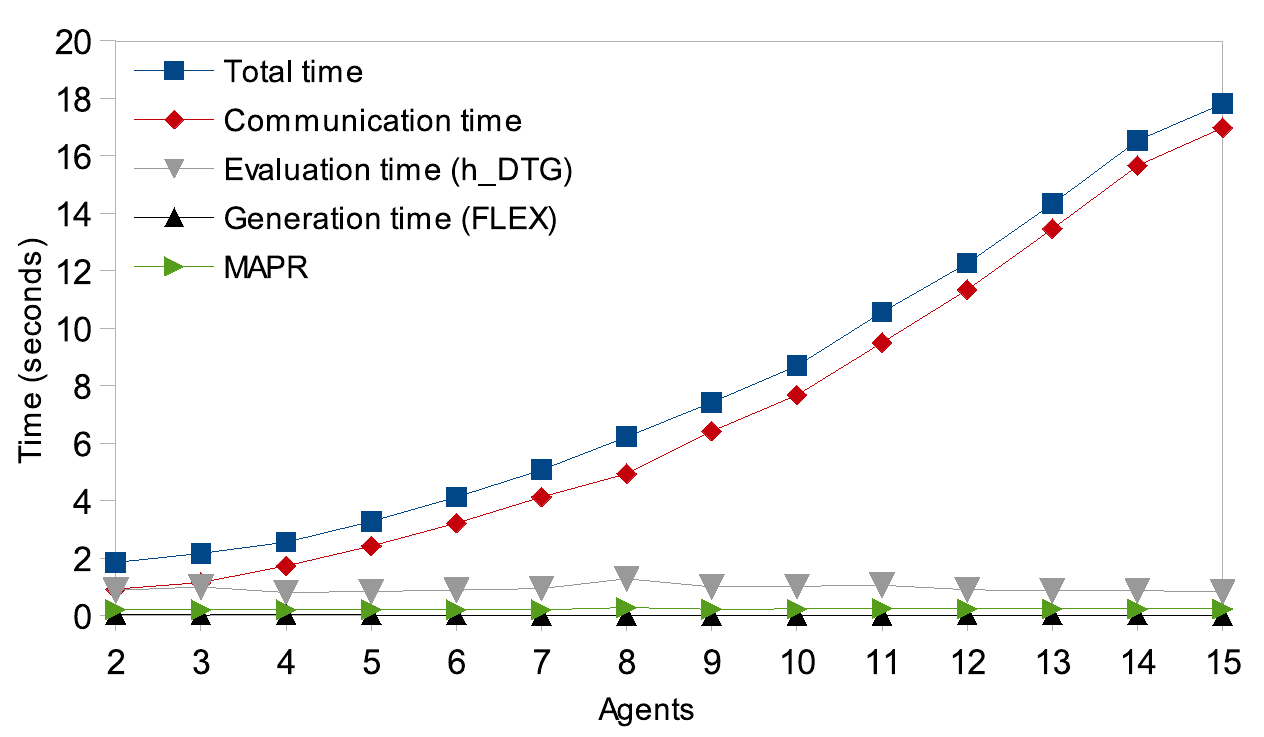}
\caption{Scalability results for the \emph{logistics}-like task}
\label{scale}
\end{figure}

The plot in Fig. \ref{scale} separately depicts the time required by each process in FMAP. We show the time required by FLEX to generate the refinement plans, the time consumed by the $h_{DTG}$ evaluation procedure, and the time spent by agents to communicate and synchronize, which includes the base plan selection and the exchange of plans among agents. Every task was solved by FMAP in 14 iterations, resulting in a 12-action solution plan (truck $t1$ and truck $t2$ each introduced six actions). 

As Fig. \ref{scale} shows, FLEX has a noticeably low impact on the overall execution time. This proves that, even when dealing with privacy and building a tree for each potentially supportable action, FLEX offers good performance and does not limit FMAP's scalability. 

Even though each task only required 14 iterations to be solved, the growing number of agents increases the size of the search tree. In the two-agent task, the agents generate an average of 3.3 refinement plans per iteration, while in the 15-agent task, the average branching factor goes up to 11.8 refinement plans. Nevertheless, this does not affect the time consumed by $h_{DTG}$, which remains relatively constant in all tasks. Since agents evaluate plans simultaneously, the evaluation time hardly grows when the number of participants increases.

Fig. \ref{scale} confirms that communications among agents are the major bottleneck of FMAP. As the number of agents increases, so does the branching factor. Therefore, each agent has to communicate more refinement plans to a higher number of participants. Synchronizing a larger number of agents is also more complex, which increases the number of exchanged messages. All these communications are managed by a centralized component, the QPid broker, which is negatively affected by the communication overhead of the system. 

\begin{figure}
\centering
\includegraphics[width=8.5cm]{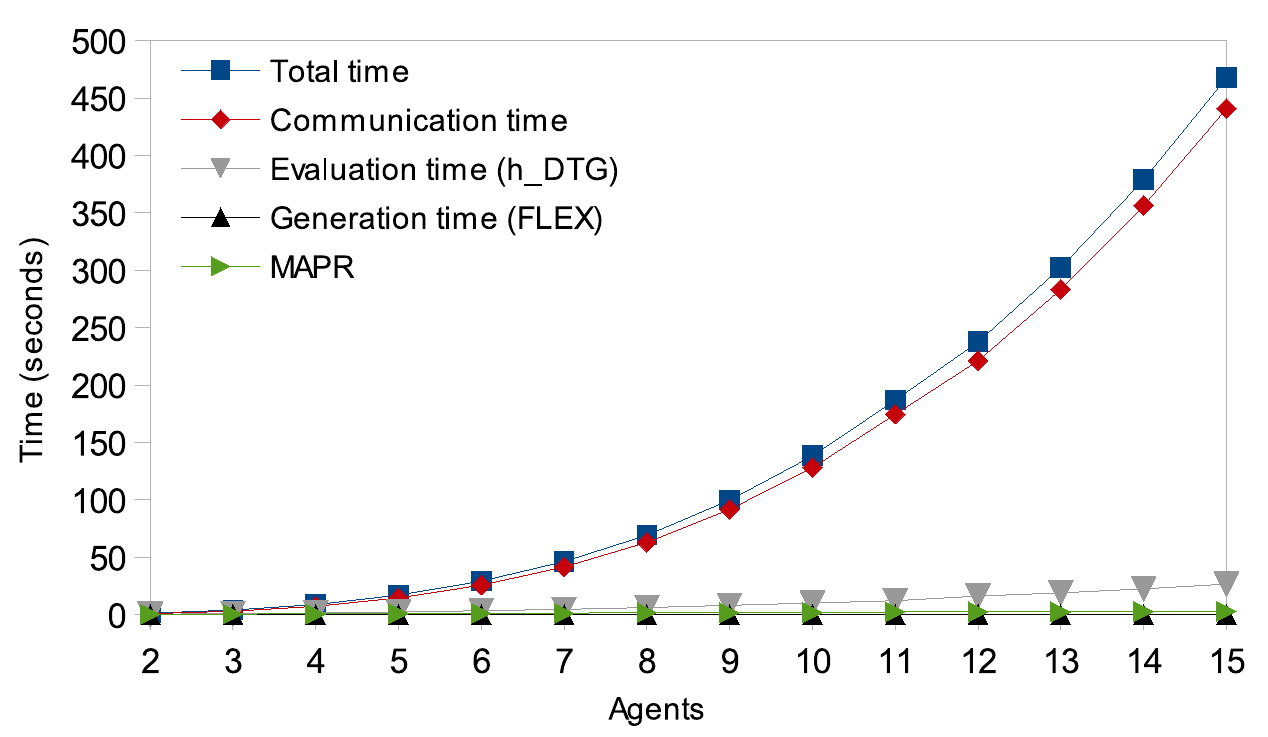}
\caption{Scalability results for the \emph{satellite} task}
\label{scasat}
\end{figure}

The behaviour of MAPR remains constant in all of the tests, taking about 0.2 seconds to resolve each task. Since MAPR does not require communications, the growing number of agents does not affect its performance. Note that if we consider only the time spent by $h_{DTG}$ (around 0.8 seconds per test) and FLEX (approximately 0.02 seconds), FMAP execution times are quite similar to MAPR. 

The resolution of this loosely-coupled task does not require coordination in order to be able to compare with MAPR. However, the coordination mechanism and message exchange of FMAP is equally applied to all planning tasks. Hence, the ability to solve tightly-coupled tasks requires great coordination, which is not the case for MAPR. 

\vspace{0.15cm}

We performed a second experiment based on the \emph{satellite} domain to assess the scalability of the two planners when both the number of agents and the number of goals increase, thus increasing the complexity of the task. We also defined 14 MAP tasks, ranging from 2 to 15 satellite agents. The simplest task comprises two satellite agents, $s1$ and $s2$, which must take an image of two different planets. The satellites are configured so that each one of them can capture an image of a single planet. The instruments they have on board are turned on and calibrated, so the agent can directly reorient and acquire the image. Unlike the first test, each \emph{satellite} task adds one more goal over the previous task, as well as an extra agent. Then, the additional agents, $s3\ldots s15$, must each solve a goal by themselves. This increases the branching factor as well as the number of iterations for solving a task. 

Fig. \ref{scasat} shows the results for this scenario. The solution plans obtained by FMAP range from 4 actions (in the two-agent task) to 30 actions (in the 15-agent task). FMAP required 31 iterations to solve the 15-agent task and only 4 iterations for the two-agent task. The growing complexity also affects the average branching factor, which ranges from 25.67 to 255.06 plans. 

As Fig \ref{scasat} shows, the complexity of the tasks does not affect FLEX, which takes less than 0.2 seconds in each task. In general, the performance of FLEX only decreases when handling very large base plans in domains with many applicable actions. We can therefore conclude that FLEX is an efficient and highly scalable component of FMAP. 

With regard to the $h_{DTG}$ heuristic, evaluation times range from 0.35 seconds for the simplest task to 26.64 seconds for the most complex one. Although the evaluation time is slightly higher than the generation time, we can affirm that this is a good performance considering that: 1) the branching factor and the number of iterations increase from task to task, which results in a much larger number of plans to evaluate; and 2) unlike FLEX, the evaluation function $h_{DTG}$ also involves some communications among agents, which obviously increase when the number of agents goes up. All in all, and considering just the times of $h_{DTG}$ and FLEX, FMAP is only about 9 times slower in the 15-agent task than MAPR, which completes this task in 3 seconds.

In summary, both tests confirm that communication overhead is the main issue of FMAP with regard to scalability. Communicating plans and synchronizing agents are rather costly tasks, especially when dealing with complex tasks that combine a large branching factor and a high number of participating agents.

\subsection{Discussion of the results}

The experimental results support our initial claim: FMAP is a domain-independent approach that offers a good trade-off between coverage and execution times being and is able to solve any typology of MAP task. 

We compared FMAP against two different state-of-the-art MAP approaches. On the one hand, MAPR is designed as a fast MAP solver. The results show that MAPR provides excellent execution times, but its performance comes at a cost: it completely rules out tightly-coupled domains that require cooperation. Many real-world domains, such as logistics or production supply-chains, require cooperation between independent entities. Hence, non-cooperative planners for solving disjoint subtasks in which agents can effectively avoid interactions are not suitable for many real-world MAP problems. Overall, in the experiments, MAPR solves 45\% of the whole benchmark while FMAP solves 73\% of the tasks.

On the other hand, MAP-POP is a general approach that is capable of solving any type of planning task like FMAP. The approach followed by MAP-POP is clearly influenced by the use of backward-chaining POP technologies and, in particular, by the application of low-informative heuristics. This planner offers the worst results in terms of coverage and execution times, thus indicating that FMAP represents a step ahead in multi-agent cooperative planning.

With regard to the scalability tests, it has been proved that the FMAP ability to scale up is only affected by communications. While MAPR performance is unaltered when the number of agents increases, FMAP performance is affected by its heavy dependency on agent communications. These results lead us to one of our future lines of work, studying techniques to reduce overhead communication without losing the ability to tackle any kind of MAP task.

\section{Conclusions}
\label{conclusions}

FMAP is a general-purpose MAP model that supports inherently distributed domains and defines an advanced notion of privacy. Agents in FMAP use an internal POP procedure to calculate all possible ways to refine a plan, which guarantees FMAP completeness. Agents exchange plans and their evaluations by means of a communication mechanism that is governed by a coordinator agent. FMAP exploits the structure of distributed state-independent domain transition graphs for the heuristic evaluation of the plans, thus avoiding having to recalculate estimates in each node of the POP search tree.

Privacy is maintained throughout the entire search process. Agents only communicate the relevant information they share with the rest of the agents. This advanced notion of privacy is very useful for modeling real-world problems. The experiments show that dealing with privacy has a relatively low impact on the overall performance of FMAP.

The exhaustive testing on IPC benchmarks shows that FMAP outperforms other state-of-the-art MAP frameworks because it is capable of solving tightly-coupled domains with specialized agents and cooperative goals as well as loosely-coupled problems. The performance of FMAP is only affected by the extensive communications among agents. To the best of our knowledge, FMAP is currently likely to be the most competitive domain-independent cooperative MAP system.

\begin{acknowledgements}
This work has been partly supported by the Spanish MICINN under projects Consolider Ingenio 2010 CSD2007-00022 and TIN2011-27652-C03-01, the Valencian Prometeo project II/2013/019, and the FPI-UPV scholarship granted to the third author by the Universitat Polit\`ecnica de Val\`encia. 
\end{acknowledgements}

\bibliographystyle{spmpsci}      

\begin{thebibliography}{10}
\providecommand{\url}[1]{{#1}}
\providecommand{\urlprefix}{URL }
\expandafter\ifx\csname urlstyle\endcsname\relax
  \providecommand{\doi}[1]{DOI~\discretionary{}{}{}#1}\else
  \providecommand{\doi}{DOI~\discretionary{}{}{}\begingroup
  \urlstyle{rm}\Url}\fi

\bibitem{Benton12}
Benton, J., Coles, A., Coles, A.: Temporal planning with preferences and
  time-dependent continuous costs.
\newblock In: Proceedings of the 22nd International Conference on Automated
  Planning and Scheduling (ICAPS), pp. 2--10 (2012)

\bibitem{Borrajo13}
Borrajo, D.: Multi-agent planning by plan reuse.
\newblock In: Proceedings of the 12th International Conference on Autonomous
  Agents and Multi-agent Systems (AAMAS), pp. 1141--1142 (2013)

\bibitem{Boutilier01}
Boutilier, C., Brafman, R.: Partial-order planning with concurrent interacting
  actions.
\newblock Journal of Artificial Intelligence Research \textbf{14}(105), 136
  (2001)

\bibitem{Brafman08}
Brafman, R., Domshlak, C.: From one to many: Planning for loosely coupled
  multi-agent systems.
\newblock In: Proceedings of the 18th International Conference on Automated
  Planning and Scheduling (ICAPS), pp. 28--35 (2008)

\bibitem{Brenner09}
Brenner, M., Nebel, B.: Continual planning and acting in dynamic multiagent
  environments.
\newblock Journal of Autonomous Agents and Multiagent Systems \textbf{19}(3),
  297--331 (2009)

\bibitem{Bresina02}
Bresina, J., Dearden, R., Meuleau, N., Ramakrishnan, S., Smith, D., Washington,
  R.: Planning under continuous time and resource uncertainty: A challenge for
  {AI}.
\newblock In: Proceedings of the 18th Conference on Uncertainty in Artificial
  Intelligence, pp. 77--84 (2002)

\bibitem{Cox09}
Cox, J., Durfee, E.: Efficient and distributable methods for solving the
  multiagent plan coordination problem.
\newblock Multiagent and Grid Systems \textbf{5}(4), 373--408 (2009)

\bibitem{Crosby13}
Crosby, M., Rovatsos, M., Petrick, R.: Automated agent decomposition for
  classical planning.
\newblock In: Proceedings of the 23rd International Conference on Automated
  Planning and Scheduling (ICAPS), pp. 46--54 (2013)

\bibitem{Dimopoulos12}
Dimopoulos, Y., Hashmi, M.A., Moraitis, P.: $\mu$-satplan: Multi-agent planning
  as satisfiability.
\newblock Knowledge-Based Systems \textbf{29}, 54--62 (2012)

\bibitem{Fikes71}
Fikes, R., Nilsson, N.: {STRIPS}: A new approach to the application of theorem
  proving to problem solving.
\newblock Artificial Intelligence \textbf{2}(3), 189--208 (1971)

\bibitem{Gerevini09}
Gerevini, A., Haslum, P., Long, D., Saetti, A., Dimopoulos, Y.: Deterministic
  planning in the fifth {I}nternational {P}lanning {C}ompetition: {PDDL}3 and
  experimental evaluation of the planners.
\newblock Artificial Intelligence \textbf{173}(5-6), 619--668 (2009)

\bibitem{Ghallab04}
Ghallab, M., Nau, D., Traverso, P.: Automated Planning. Theory and Practice.
\newblock Morgan Kaufmann (2004)

\bibitem{Gunay13}
G\"unay, A., Yolum, P.: Constraint satisfaction as a tool for modeling and
  checking feasibility of multiagent commitments.
\newblock Applied Intelligence \textbf{39}(3), 489--509 (2013)

\bibitem{Helmert04}
Helmert, M.: A planning heuristic based on causal graph analysis.
\newblock Proceedings of ICAPS pp. 161--170 (2004)

\bibitem{Hoffmann01}
Hoffmann, J., Nebel, B.: The {FF} planning system: Fast planning generation
  through heuristic search.
\newblock Journal of Artificial Intelligence Research \textbf{14}, 253--302
  (2001)

\bibitem{Jannach13}
Jannach, D., Zanker, M.: Modeling and solving distributed configuration
  problems: A {CSP}-based approach.
\newblock IEEE Transactions on Knowledge and Data Engineering \textbf{25}(3),
  603--618 (2013)

\bibitem{JonssonR11}
Jonsson, A., Rovatsos, M.: Scaling up multiagent planning: A best-response
  approach.
\newblock In: Proceedings of the 21st International Conference on Automated
  Planning and Scheduling (ICAPS), pp. 114--121. AAAI (2011)

\bibitem{Kala14}
Kala, R., Warwick, K.: Dynamic distributed lanes: motion planning for multiple
  autonomous vehicles.
\newblock Applied Intelligence pp. 1--22 (2014)

\bibitem{Koehler02}
Koehler, J., Ottiger, D.: An {AI}-based approach to destination control in
  elevators.
\newblock AI Magazine \textbf{23}(3), 59--78 (2002)

\bibitem{Kovacs11}
Kovacs, D.L.: {Complete BNF description of PDDL3.1}.
\newblock Tech. rep. (2011)

\bibitem{Krogt09}
van~der Krogt, R.: Quantifying privacy in multiagent planning.
\newblock Multiagent and Grid Systems \textbf{5}(4), 451--469 (2009)

\bibitem{Kvarnstrom11}
Kvarnstr{\"o}m, J.: Planning for loosely coupled agents using partial order
  forward-chaining.
\newblock In: Proceedings of the 21st International Conference on Automated
  Planning and Scheduling (ICAPS), pp. 138--145. AAAI (2011)

\bibitem{Lesser04}
Lesser, V., Decker, K., Wagner, T., Carver, N., Garvey, A., Horling, B.,
  Neiman, D., Podorozhny, R., Prasad, M., Raja, A., et~al.: Evolution of the
  {GPGP}/{TAEMS} domain-independent coordination framework.
\newblock Autonomous agents and multi-agent systems \textbf{9}(1-2), 87--143
  (2004)

\bibitem{Long03}
Long, D., Fox, M.: The 3rd {I}nternational {P}lanning {C}ompetition: results
  and analysis.
\newblock Journal of Artificial Intelligence Research \textbf{20}, 1--59 (2003)

\bibitem{Nissim10}
Nissim, R., Brafman, R., Domshlak, C.: A general, fully distributed multi-agent
  planning algorithm.
\newblock In: Proceedings of the 9th International Conference on Autonomous
  Agents and Multiagent Systems (AAMAS), pp. 1323--1330 (2010)

\bibitem{Obrien98}
O'Brien, P., Nicol, R.: Fipa - towards a standard for software agents.
\newblock BT Technology Journal \textbf{16}(3), 51--59 (1998)

\bibitem{Ozturk10}
\"Ozt\"urk, P., Rossland, K., Gundersen, O.: A multiagent framework for
  coordinated parallel problem solving.
\newblock Applied Intelligence \textbf{33}(2), 132--143 (2010)

\bibitem{Pal13}
Pal, A., Tiwari, R., Shukla, A.: Communication constraints multi-agent
  territory exploration task.
\newblock Applied Intelligence \textbf{38}(3), 357--383 (2013)

\bibitem{Richter10}
Richter, S., Westphal, M.: The {LAMA} planner: Guiding cost-based anytime
  planning with landmarks.
\newblock Journal of Artificial Intelligence Research \textbf{39}(1), 127--177
  (2010)

\bibitem{Rosa13}
de~la Rosa, T., Garc\'i­a-Olaya, A., Borrajo, D.: A case-based approach to
  heuristic planning.
\newblock Applied Intelligence \textbf{39}(1), 184--201 (2013)

\bibitem{Sapena08b}
Sapena, O., Onaindia, E.: Planning in highly dynamic environments: an anytime
  approach for planning under time constraints.
\newblock Applied Intelligence \textbf{29}(1), 90--109 (2008)

\bibitem{Sapena08}
Sapena, O., Onaindia, E., Garrido, A., Arang{\'u}, M.: A distributed {CSP}
  approach for collaborative planning systems.
\newblock Engineering Applications of Artificial Intelligence \textbf{21}(5),
  698--709 (2008)

\bibitem{Such13}
Serrano, E., Such, J., Bot\'ia, J., Garc\'ia-Fornes, A.: Strategies for
  avoiding preference profiling in agent-based e-commerce environments.
\newblock Applied Intelligence pp. 1--16 (2013)

\bibitem{Smith00}
Smith, D., Frank, J., J{\'o}nsson, A.: Bridging the gap between planning and
  scheduling.
\newblock Knowledge Engineering Review \textbf{15}(1), 47--83 (2000)

\bibitem{Such12}
Such, J., Garc{\'\i}a-Fornes, A., Espinosa, A., Bellver, J.: Magentix2: A
  privacy-enhancing agent platform.
\newblock Engineering Applications of Artificial Intelligence pp. 96--109
  (2012)

\bibitem{ToninoBWW02}
Tonino, H., Bos, A., de~Weerdt, M., Witteveen, C.: Plan coordination by
  revision in collective agent based systems.
\newblock Artificial Intelligence \textbf{142}(2), 121--145 (2002)

\bibitem{Torreno12ECAI}
Torre\~no, A., Onaindia, E., Sapena, O.: An approach to multi-agent planning
  with incomplete information.
\newblock In: Proceedings of the 20th European Conference on Artificial
  Intelligence (ECAI 2012), vol. 242, pp. 762--767. IOS Press (2012)

\bibitem{Torreno12KAIS}
Torre\~no, A., Onaindia, E., Sapena, O.: {A} flexible coupling approach to
  multi-agent planning under incomplete information.
\newblock Knowledge and Information Systems \textbf{38}(1), 141--178 (2014)

\bibitem{Krogt05b}
Van Der~Krogt, R., De~Weerdt, M.: Plan repair as an extension of planning.
\newblock In: Proceedings of the 15th International Conference on Automated
  Planning and Scheduling (ICAPS), pp. 161--170 (2005)

\bibitem{deWeerdt09}
de~Weerdt, M., Clement, B.: Introduction to planning in multiagent systems.
\newblock Multiagent and Grid Systems \textbf{5}(4), 345--355 (2009)

\bibitem{Yokoo98}
Yokoo, M., Durfee, E., Ishida, T., Kuwabara, K.: The distributed constraint
  satisfaction problem: Formalization and algorithms.
\newblock IEEE Transactions on Knowledge and Data Engineering \textbf{10}(5),
  673--685 (1998)

\bibitem{Feng07}
Zhang, J., Nguyen, X., Kowalczyk, R.: Graph-based multi-agent replanning
  algorithm.
\newblock In: Proceedings of the 6th Conference on Autonomous Agents and
  Multiagent Systems (AAMAS), pp. 798--805 (2007)

\end{thebibliography}


\end{document}